\documentclass{article}

    \PassOptionsToPackage{numbers, compress}{natbib}

\usepackage[preprint]{neurips_data_2022}
\usepackage{xr}
\makeatletter
\newcommand*{\addFileDependency}[1]{
  \typeout{(#1)}
  \@addtofilelist{#1}
  \IfFileExists{#1}{}{\typeout{No file #1.}}
}
\makeatother

\newcommand*{\myexternaldocument}[1]{
    \externaldocument{#1}
    \addFileDependency{#1.tex}
    \addFileDependency{#1.aux}
}

\myexternaldocument{SI}

\usepackage[utf8]{inputenc} 
\usepackage[T1]{fontenc}    
\usepackage{hyperref}       
\usepackage{url}            
\usepackage{booktabs}       
\usepackage{amsfonts}       
\usepackage{nicefrac}       
\usepackage{microtype}      
\usepackage{xcolor}         
\usepackage{graphicx}
\usepackage{multirow}
\graphicspath{ {./}}
\usepackage{algpseudocode}
\usepackage{amssymb}
\usepackage{pifont}
\usepackage[linesnumbered,ruled,vlined]{algorithm2e}
\usepackage{nomencl}
\usepackage{xcolor}

\AtBeginDocument{}
\usepackage[journal=rsc,tracking=bpchem,varioref=false]{chemstyle}

\usepackage{floatrow}
\usepackage{amsmath}
\usepackage{cleveref}

\title{Intelligent Knee Sleeves: A Real-time Multimodal Dataset for 3D Lower Body Motion Estimation Using Smart Textile}

\author{
  \textbf{Wenwen Zhang$^{1}$\thanks{Corresponding authors},~ Arvin Tashakori$^{12}$, ~Zenan Jiang$^{12}$,Amir Servati$^{2}$,~Harishkumar Narayana$^{2}$,} \\ 
  \textbf{~Saeid Soltanian$^{2}$, Rou Yi Yeap$^{2}$, Meng Han Ma$^{2}$, Lauren Toy$^{2}$, ~Peyman Servati$^{12}$\footnotemark[1]} \\
  $^{1}$Department of Electrical and Computer Engineering, University of British Columbia \\ 
  $^{2}$Texavie Technologies Inc. \\
  \texttt{\{wenwenzhang, arvin, jiang, peymans\}@ece.ubc.ca
} \\
  \texttt{\{aservati, harishkumar, ssoltanian, ryeap, meganma, ltoy\}@texavie.com} 
}

\begin{document}

\maketitle

\begin{abstract}
The kinematics of human movements and locomotion are closely linked to the activation and contractions of muscles. To investigate this, we present a multimodal dataset with benchmarks collected using a novel pair of Intelligent Knee Sleeves (Texavie MarsWear Knee Sleeves) for human pose estimation. Our system utilizes synchronized datasets that comprise time-series data from the Knee Sleeves and the corresponding ground truth labels from visualized motion capture camera system. We employ these to generate 3D human models solely based on the wearable data of individuals performing different activities. We demonstrate the effectiveness of this camera-free system and machine learning algorithms in the assessment of various movements and exercises, including extension to unseen exercises and individuals. The results show an average error of 7.21 degrees across all eight lower body joints when compared to the ground truth, indicating the effectiveness and reliability of the Knee Sleeve system for the prediction of different lower body joints beyond knees. The results enable human pose estimation in a seamless manner without being limited by visual occlusion or the field of view of cameras. Our results show the potential of multimodal wearable sensing in a variety of applications from home fitness to sports, healthcare, and physical rehabilitation focusing on pose and movement estimation.
\end{abstract}

\section{Introduction}
\label{sec:intro}

Attributed to the widespread adoption of machine learning (ML) methods in various domains, the field of computer vision has witnessed remarkable progress in the area of pose estimation \cite{zheng2020deep}. These achievements, in turn, facilitate the development of activity recognition \cite{li2018pose, singh2018eye}, point-to-point healthcare applications \cite{zhou2020single, meng2020wireless,ahad2019vision}, augmented reality (AR) \cite{lei2019survey}, and human-computer interactions \cite{maurice2019human}. Images and videos are usually the main sources for ML models to extract human pose, with major challenges including multi-person pose estimation, occlusion, and limited field of view (FoV) of cameras \cite{Chen_2018_CVPR}. Moreover, concerns for data privacy in camera-based methods also encourage non-vision-based frameworks \cite{zhao2018rf, zhao2018through} for human pose estimation that can provide more private data gathering. Since human motion must involve muscle activation, stretching, and contraction, we propose a pair of Smart Knee Sleeves with embedded yarn stretch sensors and Inertial Measurement Units (IMUs) to detect muscle contractions and joint movements, reflecting human movements. Recent advances in flexible electronics have demonstrated the feasibility of advanced wearable sensor motion capture (MoCap) and pose estimation \cite{luo2021intelligent, luo2021learning, delpreto2022actionsense, shi2020deep} with different form factors and performance parameters. 
Closing the gap between the current portable wearable devices' ability to estimate human posture and more accurate joint angle and movement estimation holds immense potential for facilitating healthcare applications, aiding individuals with joint-related illnesses (such as arthritis, rheumatism, or osteoporosis), as well as assisting in sports analysis \cite{ghosh2023sports, seshadri2019wearable}.

\begin{figure}[t]
  \caption{\textbf{Overall outline of the intelligent Texavie MarsWear knee Sleeves based 3D pose estimation process including the data collection, hardware setup, and qualitative results.} (a) Marker-based camera setup to capture major joint angles of the lower body during the exercises. The output time-series data recording joint movements will be used as supervised annotations in training steps. \ding{172}-\ding{177}: MoCap cameras; \ding{178}: subject location for data acquisition.~(b-c) Photographs of the experimental environment during data collection incorporating the wearable sensors. 
  (d) An unfolded version of Texavie MarsWear Smart Knee Sleeve, displaying the location of the PCB hardware box, removable battery box, Bluetooth connection, stretchable interconnects, pressure sensors, and IMUs. (e) Major joints included in the training and testing process. (f) Visualization of the 3D human model for lower body pose estimation for both the MoCap camera system and smart Knee Sleeves. (g) Schematic of smart Knee Sleeves work by a user.}
  \label{fig:fig0}
  {\includegraphics[width=1\linewidth]{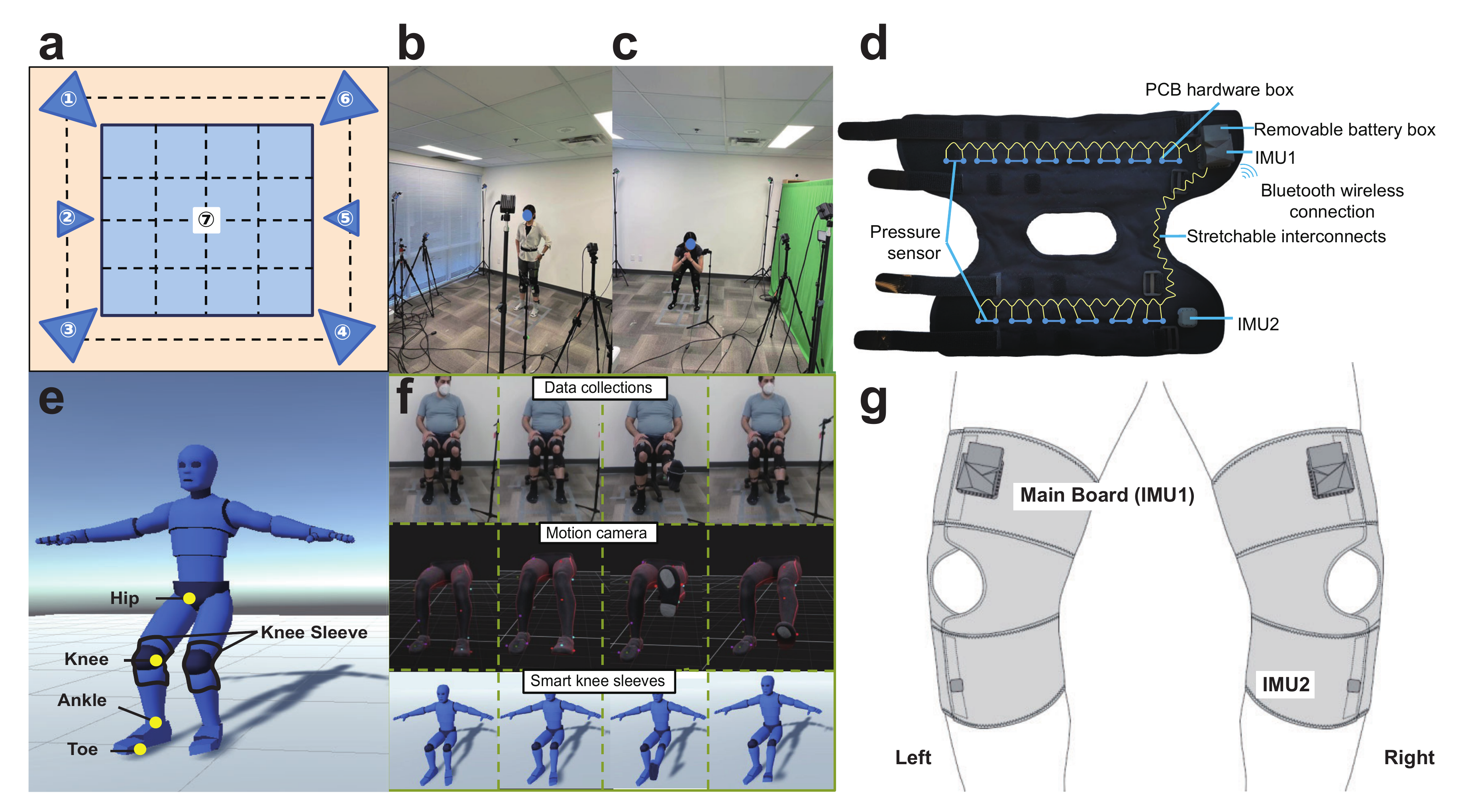}}
\end{figure}
In this research, we introduce a comprehensive dataset with extensive ground truth labels from MoCap camera systems and a baseline model for pose estimation tasks based on an overview architecture as displayed in \autoref{fig:fig0}. Here, \autoref{fig:fig0} (a) depicts the camera setup for our MoCap system, which provides the ground truth labels for our supervised learning. \autoref{fig:fig0} (b-c) show the data collection process displaying the relative position of the subject wearing the Knee Sleeve device and cameras. Smart Knee Sleeves (provided by Texavie) are crafted from stretchable and washable textile materials (Weft knitted double-jersey rib fabrics composed of polyester/spandex) as shown in \autoref{fig:fig0}(d) and (g). The smart textile device is embedded with yarn-shape pressure sensors located around the hamstring and quad muscles as well as calf and shin muscles on the legs of the user and two IMUs above and below the knee joints. Wavy 3D stretchable interconnects connect all sensors and IMUs to a wireless readout and processing board with a rechargeable battery. 

\begin{figure}[t]
  \caption{\textbf{Architecture of the 3D pose estimation machine learning (ML) model.} The baseline ML model architecture utilized in this work to estimate joint movements. The input is sensor signal readouts with a sliding window from our smart Knee Sleeves, and the output is the joint motion in quaternion. We visualize the output quaternions through a Unity3D human model. }
  \label{fig:fig0_1}
  {\includegraphics[width=1\linewidth]{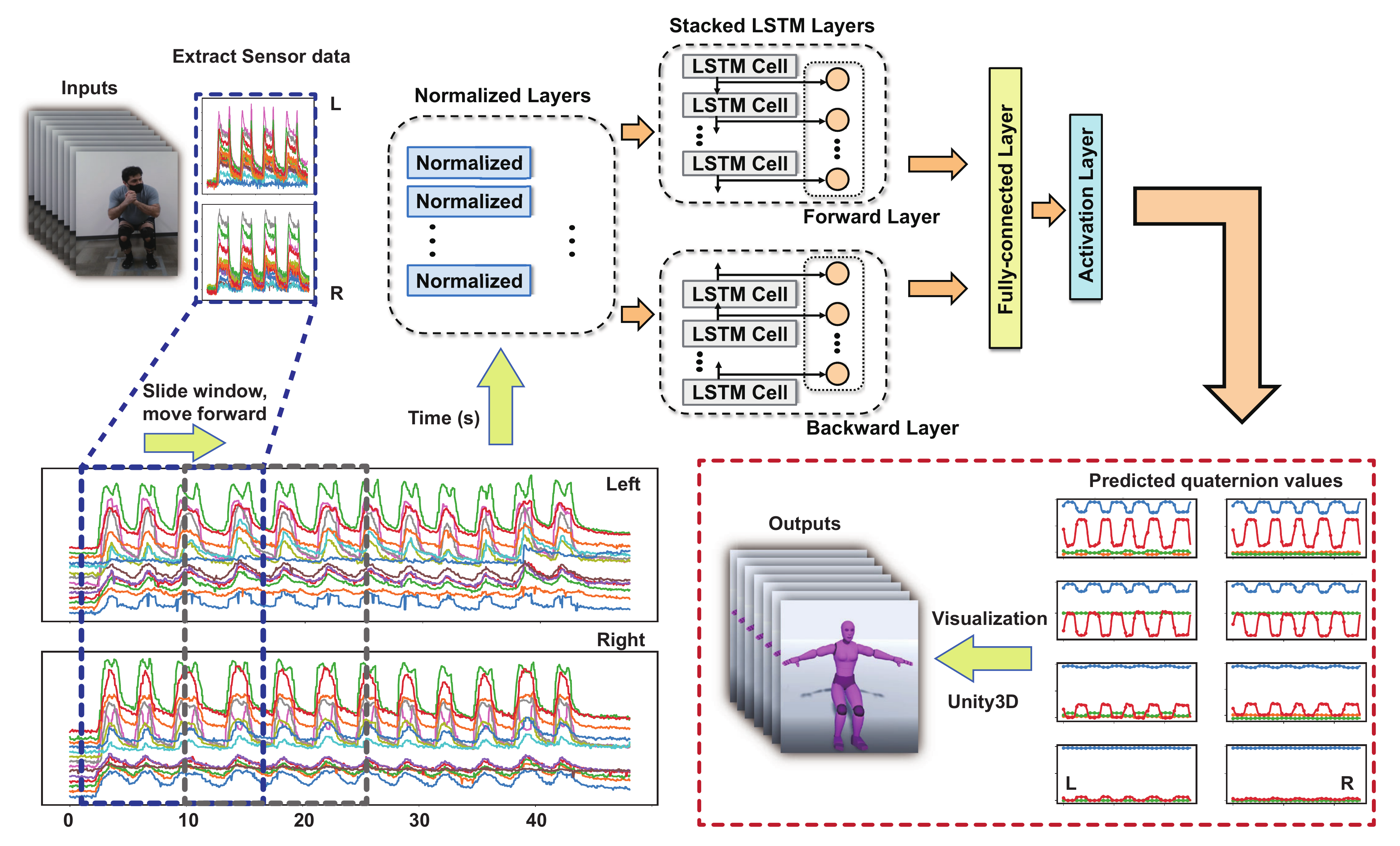}}
\end{figure}

\begin{figure}[t]
  \caption{\textbf{Qualitative Results of Smart Knee Sleeves Across Time Steps.} The depicted poses, from left to right, are squatting (a), hamstring curling (b), and leg raising (c). For each sequence, from top to bottom, we showcase our data collection setup, ground truth annotations captured by the MoCap, and the qualitative outcomes derived from our knee sleeve readouts displayed using a human model in Unity3D.}
  \label{fig:quat}
  {\includegraphics[width=1\linewidth]{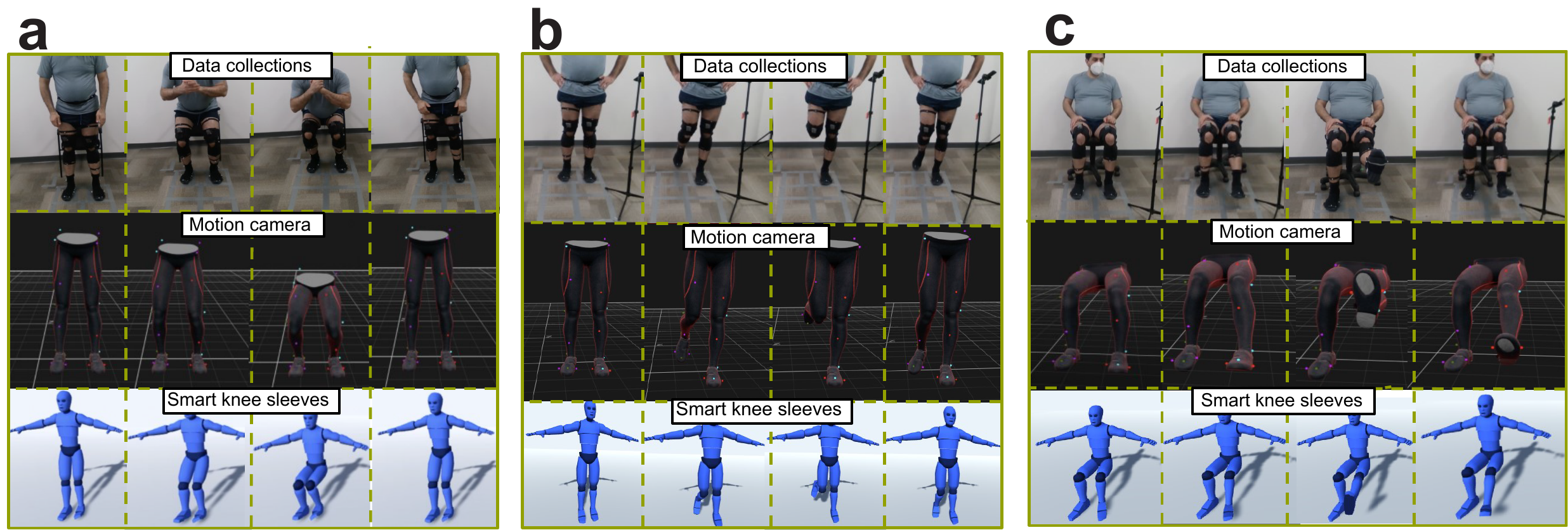}}
\end{figure}

We monitor four major joints of each leg (hip, knee, ankle, and toe) on the left and right sides separately, using MoCap system. Our Smart Knee Sleeves provide 14 channels of pressure sensor data, indicating muscle contractions related to movement, and 9 channels of IMU data that capture the angle of the knee joint. The unfolded version of our knee sleeves is schematically displayed in \autoref{fig:fig0} (d), showing the location of the PCB, embedded pressure sensors, removable battery box, IMUs, and stretchable interconnects. Our smart knee sleeve and custom-made software developed by Texavie Technologies Inc, work together through a special wireless communication system, enabling us to record real-time reactions of muscles and joints during exercise and movements as displayed in \autoref{fig:fig0} (d). Developed iOS app and supporting software to enable easy collections of various daily exercise poses from the Knee Sleeves. With the assistance of MoCap guidance, we have developed a recursive neural network-based model that can estimate 3D human lower body joint angles by fusing data from IMUs and pressure sensors, as illustrated in \autoref{fig:fig0_1}. The neural network utilizes normalized sliding-window sensor fusion data from our smart Knee Sleeves to generate real-time time-series quaternions that estimate the motion of all joints of the lower body. The 3D human model visualization is developed in Unity3D. The qualitative visualization results in \autoref{fig:quat} exhibit comparative outcomes from RGB images during data collection, ground truth quaternions extracted from the MoCap system, and estimated 3D human model visualization results from Knee Sleeves. 

Our ML model under the supervision of the commercialized MoCap system information, 
can accurately predict the 3D human pose with an average joint angle error of 7.21 degrees, compared to the ground truth data obtained from the MoCap system. Furthermore, we evaluated the model's ability under different scenarios to generalize to new individuals and poses. The proposed smart Knee Sleeves can overcome the challenges of occlusion and multiple-person detection faced by camera systems. Through the use of smart textile force/stress sensing fused with IMU data, the proposed solution opens up possibilities for human pose estimation that is unaffected by visual barriers and can be executed seamlessly and privately. To the best of our knowledge, this is the first work to propose the prediction of lower body 3D human joint angles solely from a pair of customized, stretchable, wireless smart knee sleeves. To sum up, the principal achievements of this article include:

\begin{itemize}
\item A comprehensive multimodal dataset with synchronized wearable recordings of embedded pressure sensors, IMU, and marker-based MoCap data on major joints of the lower body.
\item A baseline model on time-series data for 3D predictions of major joints on the lower body with an average of 7.21 degrees, going beyond the knee joints and to other joints using the smart Knee Sleeve. The public access to our synchronized dataset and baseline model is at \href{https://feel.ece.ubc.ca/smartkneesleeve/}{https://feel.ece.ubc.ca/smartkneesleeve/}.
\item Extension and generalization of our prediction model to unseen exercises and individuals.
\end{itemize}

The following is the organization of the paper: Initially, we provided a summary of current 2D and 3D human pose estimation techniques, followed by an investigation of proposed benchmarks for kinesthetic sensing in textile-based wearable sensors in \autoref{sec:related}. Then, we presented the specifics of our smart wearable sensor dataset in \autoref{sec:dataset}, including how we acquired and pre-processed the data. Afterward, we described the implementation details, baseline models, and performance for our dataset in \autoref{sec:implementation}. Following that, we discussed the limitations of our baseline model and analyzed their causes in \autoref{sec:limitation}. Lastly, we wrapped up the paper in \autoref{sec:conclusion} and included supplementary materials for additional information.

\section{Related Work}
\label{sec:related}
\subsection{Human Pose Estimation}

The presence of emergent human pose datasets and the introduction of deep neural network models have led to significant advancements in human pose estimation from images or videos in recent years. MoCap system are used as ground truth for these studies. As shown in \autoref{fig:fig0} (a-c), we used reliable MoCap systems (Optitrack) to provide supersized annotations in our training process. We used six cameras around the subject (marked by 7) to fully capture the motion in 3D planes as shown in \autoref{fig:fig0}(a). Calibration is required every time before data collection since the relative location between the subjects, markers, and cameras will have a great influence on the final outputs from camera-based algorithms. Key-points estimation on joints to predict human pose \cite{wang2021robust, jin2020whole, martinez2017simple, li2022mhformer, papandreou2017towards} has been a popular method in the human pose inference area. Multi-view \cite{zhang2021direct}, special data augmentation \cite{gong2021poseaug} or multi-modal data \cite{an2022mri, chen2015utd} are usually required to assist in the prediction of 3D key points with vision and camera-based methods. Subsequently, the demand to extract more detailed information about the human body's posture and movements has driven the interest in 3D pose estimation utilizing 3D human models \cite{kanazawa2018end, Osman2020STAR, loper2015smpl}. Pose estimation with 3D human models are capable of providing more details on the orientation of the body joints,  skeletal structure, etc, and thus is more resource-intensive in computer vision tasks. 

To capture more detailed information about joint angles and movements while requiring lower computational resources, wearable sensors have emerged as a promising alternative to camera-based methods for 3D pose estimation. Camera-based methods largely rely on visual cues to infer the position and orientation of body joints, and face many challenges including fixed equipment location \cite{cao2017realtime, ahad2019vision}, lighting conditions \cite{lei2019survey}, environment, background noise \cite{simo2012single, martinez2017simple}, occlusion \cite{zhang2022voxeltrack}, and multi-person problems \cite{fang2017rmpe, wang2022distribution}. Wearable sensors, on the other hand, avoid these issues as they don't require a clear and unobstructed view of the body. Meanwhile, flexible electronics can provide real-time measurements of the dynamic movement status of the human body segments and are a reliable source of kinesthetic information under a wide range of conditions, including outdoors, low-light, noisy, and cluttered environments. 

IMUs-based kinesthetic sensing \cite{von2017sparse, huang2018deep, yi2022physical,guzov2021human, jiang2022avatarposer} excel in wearable pose estimation primarily due to their self-contained operation, eliminating the need for external references or beacons. Their compact design ensures user comfort, while their capacity to integrate with other sensors, like magnetometers, boosts accuracy and mitigates drift. \cite{jiang2022avatarposer} employ IMU-based equipment positioned on the head and hands to predict comprehensive full-body poses in Mixed Reality, which overcomes the constraints of existing systems that provide only partial virtual representations. \cite{huang2018deep, yi2022physical} aim to efficiently predict precise human poses with a mere six strategically placed IMUs (XSens) on the body, addressing the complications associated with traditional dense configurations and meeting the rising needs of interactive technologies. However, using solely IMUs for pose estimation faces some essential challenges, notably the drift errors that accumulate during position calculation by velocity integration or orientation determination by angular velocity integration. Supplementary technologies such as Kalman filtering, sensor fusion with other systems, or periodic recalibration are imperative to achieve optimal accuracy.

Our research goes beyond traditional vision-based and standalone IMU methods in adeptly detecting subtle, real-time changes in joint angles and movements. The Smart Knee Sleeves integrate both IMUs and pressure sensors to reduce drift errors effectively. These sleeves are convenient and designed for everyday wear, eliminating the need for any additional equipment to monitor daily activities and exercise routines. We deliver real-time 3D human models with details on 8 major joints of the lower body, which are immensely valuable for sports therapists to provide feedback on athletes' technique, rehabilitation \cite{madadi2020smplr} for tracking the progress of patients undergoing physical therapy \cite{liao2020deep} and enhancing human-computer interaction (HCI) \cite{pantic2003toward, jaimes2007multimodal} and virtual reality (VR) experiences \cite{wen2020machine}. Those applications extend beyond the realm of mere 3D human pose estimation, benefiting various facets of society.

\begin{figure}[t]
  
  \caption{\textbf{Normalized sensor signal during different exercises.} The exercises from left to right are (a) squat, (b) hamstring curl, and (c) leg raise, respectively. The electric signals generated by the pressure yarn sensors correspond to the degree of stretching and muscle contractions. In the absence of movement in a resting leg, flat lines for the right leg for the hamstring curl (bottom sub-panel b) and leg raise (bottom sub-panel c) are shown for better comparison, where only the left leg is intentionally moving. This describes the kinematic process yielding the sensor output depicted within the illustrated diagram. }
  \label{fig:Sensor}
  {\includegraphics[width=1\linewidth]{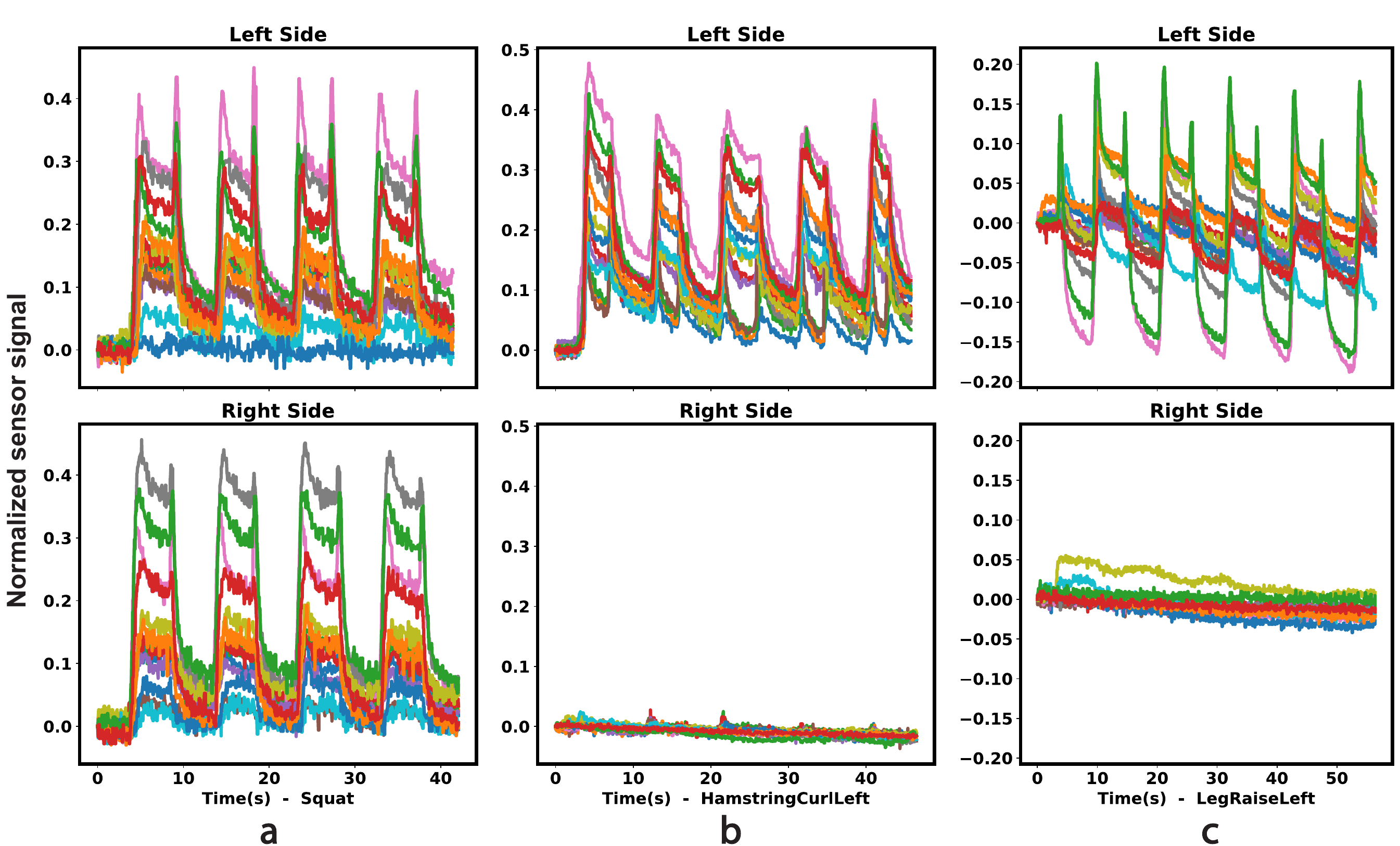}}
\end{figure}

\subsection{Kinesthetic Sensing through Smart Textile Fabric}

Advancements in stretchable smart textiles \cite{libanori2022smart, chun2021artificial} have enabled the development of wearable devices that are well-suited for dynamic tracking, monitoring, and modeling of human movements in a variety of contexts. With the ability to detect kinesthetic feedback during body movement via detecting force-induced deformations in muscle activation \cite{zarate2022computational}, stretchable smart textile modalities hold great possibilities for predicting 3D human pose with great accuracy. Designs for detecting human activities through textiles have been investigated in several major ways: producing electrical signals through human-environment contacting \cite{luo2021learning, luo2021knitui, luo2022digital, zhang2021dynamic}, pressure change \cite{zhou2020single}, and material deformation \cite{meng2020wireless}. With those characteristics, textile-based sensors have been fabricated as wearable apparel/garments to wear on diversified parts of the body such as face \cite{guo2023texonmask}, arms \cite{xu2022smart, alam2022smart}, and hands \cite{zhang2021dynamic, sundaram2019learning}, to capture the dynamic status of the designated area. Luo et al. \cite{luo2021intelligent} predicted the key points of joint angle from tactile carpet, which partially solve multi-person problems. However, their data include large-scale no interaction data, where no pressure is produced with no subject standing. Zhang et al. \cite{zhang2021dynamic} proposed e-textile gloves to sense contact with objects and movement, which also include a large portion of background data and are unable to provide joint details of the finger. Xu et al. \cite{xu2022smart} adopt responsive pressure sensors to detect arm movement, but they focus on classification tasks only. 

Distinguished from previous work focusing on wearable sensors, which are mostly classification tasks or unable to provide direct information on joint angles and motions, we aim to predict major joint movement in the lower body with subtle details of orientation and bending in three directions as illustrated in \autoref{fig:fig0} (e,f). The pressure sensor and IMUs are designed around the thigh and calf regions to capture the kinesthetic feedback from different orientations. We provide complete pipelines from ML-based joint prediction to human 3D model reconstruction with the assistance of Unity3D as depicted in \autoref{fig:fig0_1}. 

\section{Smart Wearable E-textile Sensor Dataset}
\label{sec:dataset}

\paragraph{Data Acquisition}
Our stretchable knee sleeves include 14 channels of sensor arrays and 9 channels of IMU data from two Bosch Sensortec BNO055 IMUs. A customized readout circuit board is designed and fabricated by Texavie Technologies Inc., to arrange and fuse multiple channels of data from both pressure sensors and IMUs, and to capture subtle changes around major joints during exercise. Paired with specialized mobile software constantly communicating with the hardware through Bluetooth low energy protocol, we are able to acquire data free of wires and realize the real flexibility and wearable to track human movements. Our smart Knee Sleeves are personalized, robust, and highly reliable for data collection under various physical conditions and exercises. Under the paired Bluetooth connectivity, we acquire over 300 sensing readouts at a 20 Hz sampling rate for the left and right knees. As shown in \autoref{fig:Sensor}, our smart Knee Sleeves exhibit high responsiveness to changes in muscle contraction and relaxation during exercise poses. The pressure sensors remain stable in the absence of external stress or deformation. During exercises such as squats, hamstring curls, and leg raises, the pressure sensors on both the left and right knee generate electric signals that correspond to the level of stress sensed at designated locations. In the case of the squatting pose, the left and right knee signals are similar due to the comparable muscle reactions on both sides of the body, carrying information about the symmetry of movement and muscle forces. For the hamstring curl and leg raise poses, we observe more significant pressure sensor responses on the left side than the right side as it serves as the primary exercise leg. 

Using the multi-modal data from multiple channels,  we are capable of estimating angles for major joints of the lower body during subject's movements. We have acquired over 140,000 synchronized frames of data from our stretchable wearable smart textile modality and MoCap system from 12 continuous days from different subjects with various sizes of Knee Sleeves. The details of subject numbers, task numbers, and other details are summarized in \autoref{tab:dataset_details}. For ethical considerations, please refer to \autoref{sec:ethical}. 


\paragraph{Data Pre-processing and Augmentation}
We extract ground truth data from the MoCap system, where markers are required to calculate joint angles. We calculated the relative angles of the joints from the MoCap system and used these angles as supervised labels in the training task. The output label contains 8 joints' time-series quaternions for the left and right legs, respectively, as illustrated in the label-generation process \autoref{fig:Label_gen} of \autoref{apd:apd}. The details including the content, structure, and dimension of our dataset are summarized in \autoref{apd:data_structure}. This is an example of generating time-series labels from a squatting exercise. It is important to recognize that occlusion problems can impact MoCap systems, leading to inaccuracies (refer to \autoref{fig:motion_compare} for details) in ground truth labels. As a result, this can generate errors during subsequent training procedures. But this error is not caused by our model or wearable devices and can be alleviated by removing unreasonable ground truth annotation data. But as the baseline model, we incorporated the entirety of the MoCap system's collected data to ensure data integrity. This also verifies that wearable sensor integrative smart sleeve benchmarks are more accurate and reliable than computer vision methods under certain circumstances where occlusions occur. 


Although Bluetooth and wireless communication have contributed to the development of flexible and mobile devices for use in daily activities and exercises, the latency of Bluetooth \cite{liu2021comprehensive} may cause uneven time intervals. Similarly, we observed uneven time intervals for our smart knee sleeves as well, whereas the MoCap system consistently provides an evenly increased time axis. To align the data from the MoCap system with the output from the Knee Sleeves, we employ the Fourier method to resample the Knee Sleeve readouts. 

\begin{figure}[t]
  \caption{\textbf{Quaternion distance and estimation results comparison}. (a). The model's overall performance evaluated on the entire dataset, encompassing all exercises and individuals. (b). The quaternion output from the models. The knee angle prediction showed the highest level of accuracy across all joints. The toe angle was found to be mostly stable with minimal movement during the squat exercise. }
  \label{fig:all_seen}
  {\includegraphics[width=1\linewidth]{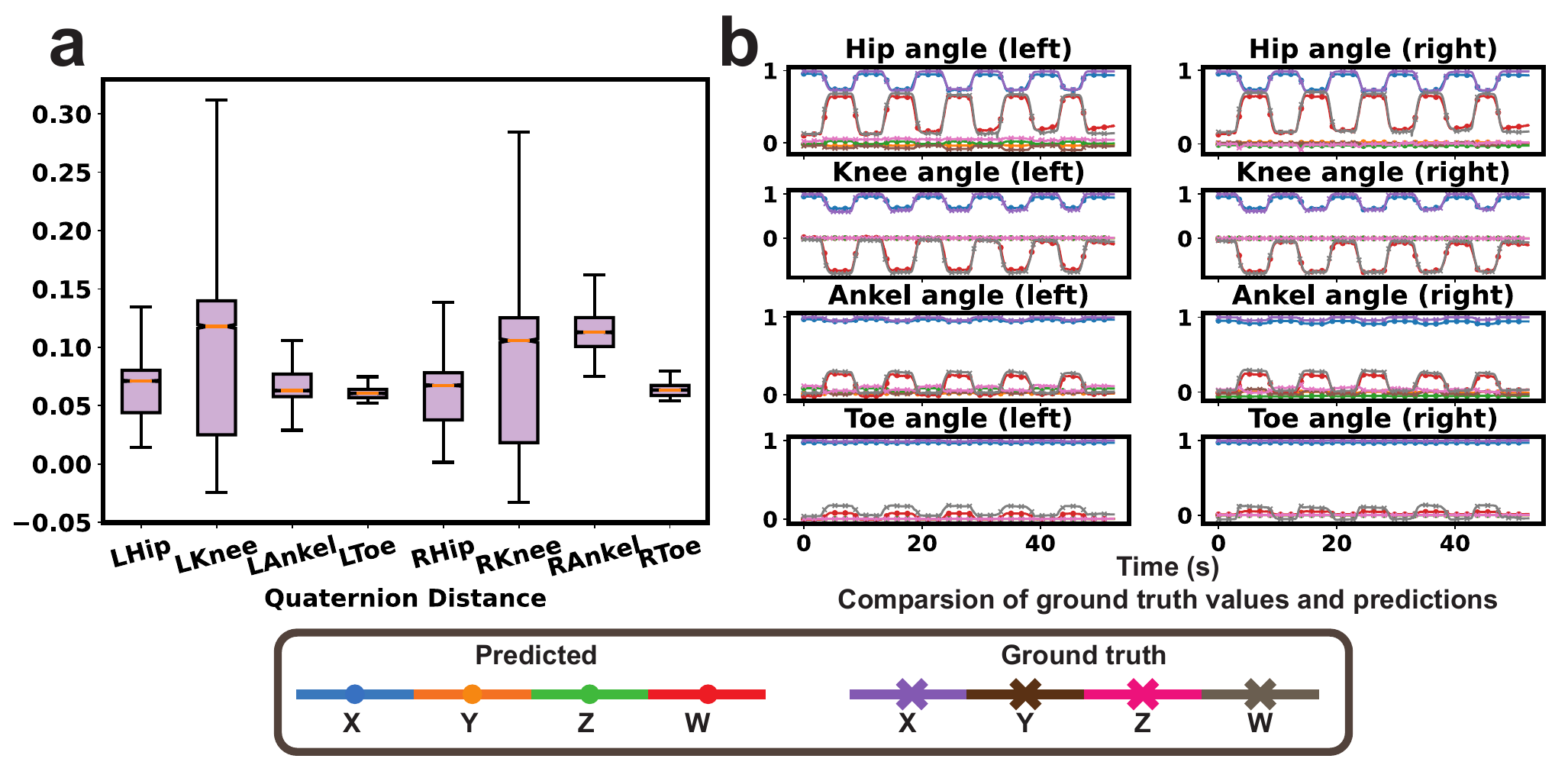}}
\end{figure}

\begin{table}[hbtp]
\centering
\caption{\textbf{RMSE in degrees for smart Knee Sleeves performance evaluation on various scenarios.} The first row is RMSE for all seen tasks, while the second to fourth rows are RMSE for unseen squats, hamstring curls, and leg raise exercises, respectively. Refer to \autoref{tab:RMSE_degree_si} for more details.}
\label{tab:RMSE_degree}
\setlength{\tabcolsep}{1.5mm}{
{\begin{tabular}{|c|c|c|c|c|c|c|c|c|c|}
\toprule
\textbf{Scene} & \textbf{Pose} & \textbf{LHip} & \textbf{LKnee} & \textbf{LAnkel} & \textbf{LToe} & \textbf{RHip} & \textbf{RKnee} & \textbf{RAnkel} & \textbf{RToe} \\   \midrule
All\_seen & Avg & 9.03 & 11.80 & 6.23 & 3.81 & 9.31 & 7.69 & 7.04 & 2.77 \\ \hline
\multirow{3}{*}{\begin{tabular}[c]{@{}c@{}}Unseen \\ Tasks\end{tabular}} & BendSquat & 17.50 & 14.20 & 12.30 & 4.25 & 17.90 & 15.10 & 12.10 & 5.12 \\ \cline{2-10} 
 & \begin{tabular}[c]{@{}c@{}}Hamstring \\ Curl\end{tabular} & 12.70 & 18.00 & 6.13 & 2.71 & 12.40 & 16.90 & 6.49 & 4.13 \\ \cline{2-10} 
 & Leg Raise & 10.20 & 19.80 & 9.05 & 2.56 & 9.55 & 16.20 & 9.29 & 5.50 \\  \hline
\end{tabular}}}
\end{table}

\section{Implementation Detail and Experimental Results}
\label{sec:implementation}

\paragraph{Implementation} We implement the baseline neural network using 2 layers of long short-term memory (LSTM) with Pytorch \cite{paszke2019pytorch}, as shown in \autoref{fig:fig0_1}. We use data from the pressure sensors and IMUs as input and that from the MoCap system as ground truth labels to train the LSTM model. The output from our ML model is the quaternions of the eight major joints of the lower body. The sequence length used to create the sliding window is 250 sample index to capture the change of pressure sensors and IMUs during movement. We choose the tanh function as activation to match the output range of quaternions from -1 to 1. All sensor readings and quaternions should be normalized to range (-1,1) before training to bring features to a similar scale. 
We ran the experiment on NVIDIA GeForce RTX 2060 and got results within 5 hours.

\paragraph{Experiments} We trained our ML model on 109,000 pairs of MoCap and wearable sensor output frames and validated and tested on 13,000 frames of data. We evaluated the results with quaternion distance ($D_q$) to compare the estimated 3D joints' angles with corresponding values from MoCap ground truth data, as shown in \autoref{fig:all_seen} (a). The calculation of $D_q$ (see \autoref{eq:Q_d} for details) is performed under the scale of normalized quaternions \cite{huynh2009metrics}.  \autoref{fig:all_seen} (b) illustrates an example output of quaternions for the squat exercise separately derived for the left and right legs. The motion recorded by the pressure sensors aligns well with the changes of quaternions, as displayed in \autoref{fig:Sensor}. \autoref{tab:RMSE_degree} summarizes the root-mean-square error (RMSE) of each joint in 
degrees converted from quaternion distance results. We report average errors of 9.16, 9.75, and 6.64 degrees for the knee, hip, and ankle angles, respectively. The toe has a relatively low margin of error because it is not a primary joint used in squats and is not significantly involved in activities during exercise.  

\begin{figure}[htbp]
  \caption{\textbf{Device's generation evaluation (unit: normalized quaternions).} \textbf{Top (a-c)}: Qualitative performance extracted from unseen tasks for squat, hamstring curl, and leg raise exercises. \textbf{Middle (d-f)}: Error of joints for predictions of seen tasks. The model sees all the data from participants and tasks. \textbf{Bottom (g-i)}: The occurrence of prediction joint quaternion error in new tasks and individuals. The training was conducted on a partitioned dataset that excluded tested actions to exam the generalization of our model to unseen tasks.}

  \label{fig:unseen}
  {\includegraphics[width=1\linewidth]{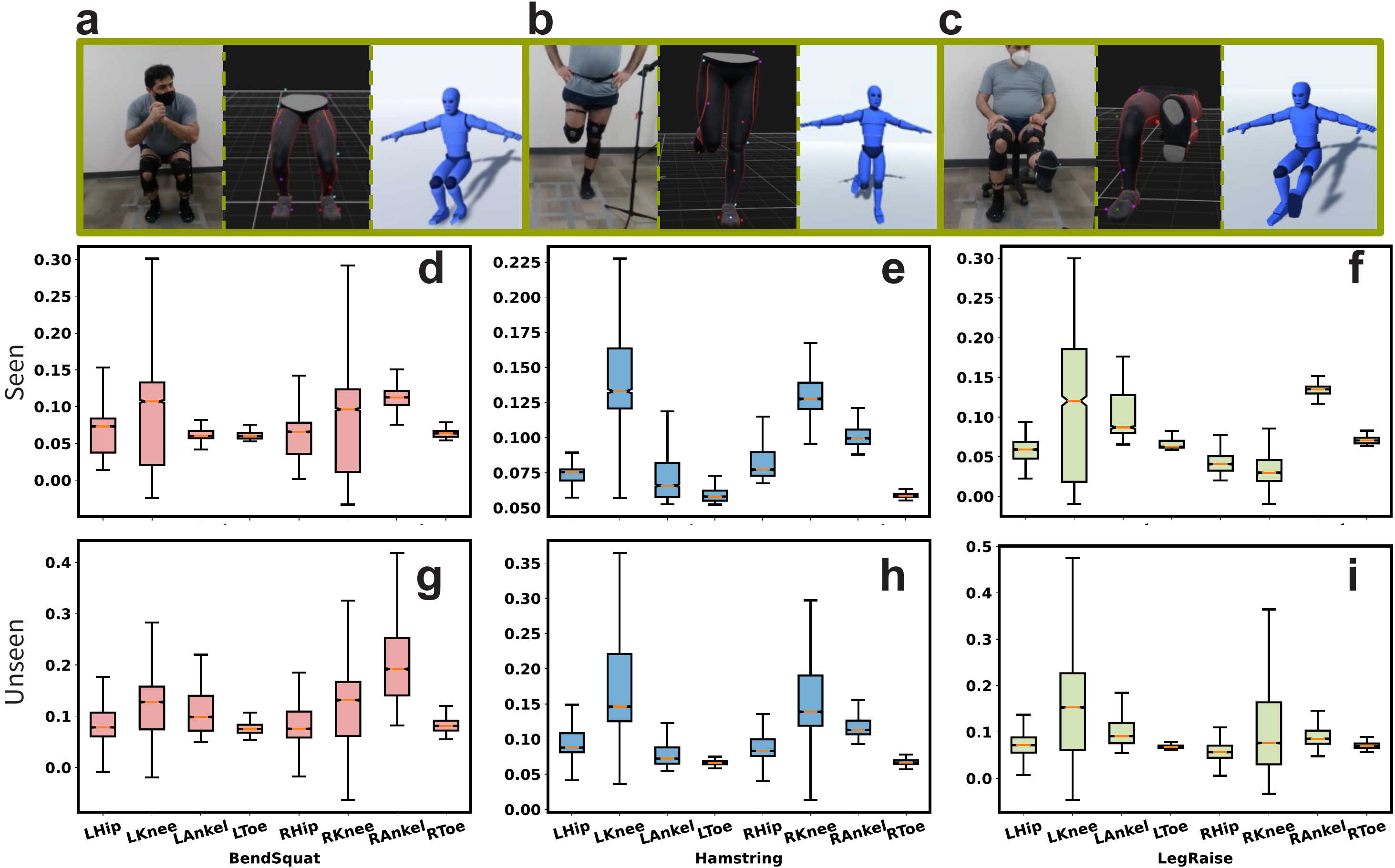}}
\end{figure}

Our assessment involves measuring the device's ability to estimate joint angles for activities that have not been previously observed. As displayed in \autoref{fig:unseen} (g-i), the model performs well on different unseen tasks. Our dataset is roughly categorized into three types of exercises: squat, hamstring curl, and leg raise.  Despite the various forms of squatting available (stepwise squat, tired squat, etc. See \autoref{tab:exercise_list} for details), we view them as the same movement when it comes to training. In our trials of unseen exercises, we exclusively evaluated the bent squat \autoref{fig:unseen}~(d, g) because other variations of squatting produced very similar results. To estimate the bent squat, we trained solely on exercises involving leg raises and hamstring curls, excluding all other types of squats from the training process. In theory, regardless of the type of exercise performed, the pressure sensor and IMUs should exhibit comparable patterns as long as there is similar muscle contraction and extension since the human pose is essentially linked to muscle activation. 

Our smart knee sleeve generalizes to unseen poses with slightly increased errors as for the case of hamstring curls \autoref{fig:unseen} (e, h). The reasonable degradation of performance in unseen tasks can arise from the mildly distinctive patterns in leg raise. Leg raise in \autoref{fig:unseen} (f, i) is the only pose in our dataset that starts from a sitting position. Since we are measuring the pressure sensor and IMUs with relative values to avoid the sensor and marker displacement influence, the start point for both strain sensors and IMU data is zero. This is reasonable for the poses that start with the standing position. However, for the sitting position, the supervised labels provided by the MoCap are actually 90 degrees, and the pressure sensors will also have initial values with stress applied. To eliminate the effect, we rotate all the quaternions of leg raise from IMUs 90 degrees before training. The pressure sensor data will also have a relatively influenced pattern due to the initial sitting position. The inconsistency between IMUs, sensors, and ground truth data induces confusion and errors in the model inference process. If we let the model see only 10\% of the leg raise data in the training process, the performance will be improved with less error (see \autoref{fig:legRaise_compare} of \autoref{apd:apd}). 

\begin{figure}[htbp]
  \caption{ \textbf{Quaternion distance for unknown individual exercises (a) and unseen dates (b).} The model's performance remains consistent when trained with unknown individuals and dates, with only a minimal rise in the quaternion distance error, indicating its strong generalization capabilities.}
  \label{fig:unseen_people}
  {\includegraphics[width=1\linewidth]{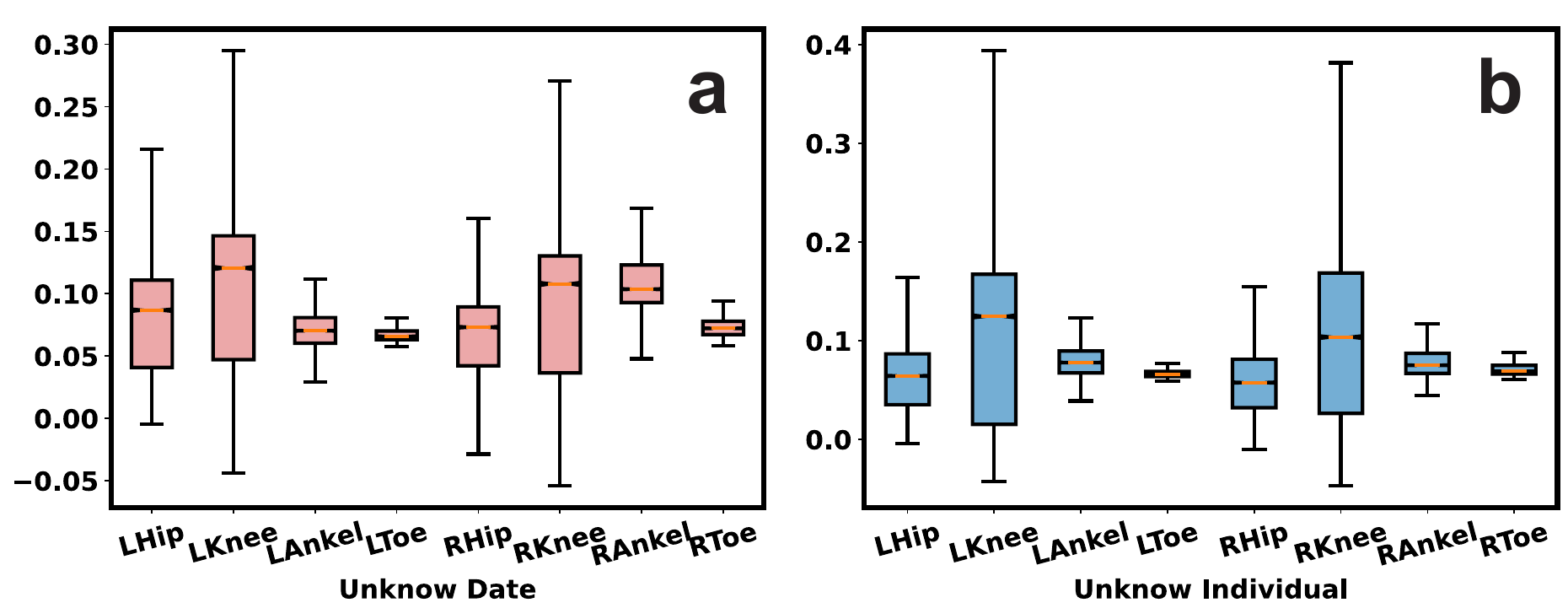}}
\end{figure}

Our wearable Knee Sleeves are customized to fit each individual perfectly. Except for poses that have not yet been encountered, it is possible that the wearable electronics, markers, and MoCap system calibration positions may shift when tested on different dates and individuals. We have conducted tests under these conditions and have depicted the results in \autoref{fig:unseen_people}. No discernible decrease is observed in the outcome and accuracy of the model. 

\section{Discussions and Limitations}
\label{sec:limitation}
Our attempts to obtain precise angle measurements from the lower body's anatomical pose have encountered multiple challenges that can compromise measurement accuracy during exercises. These challenges include soft tissue movement and sensor displacement during prolonged exercises, which can result in potential errors. Moreover, the accuracy of our model inference is compromised when testing the leg raise pose, which is the only pose starting from a seated position. To overcome these challenges and enhance our system, we will include additional scenarios that involve transitioning from sitting to standing or lying down to examine the impact of the starting position on our smart Knee Sleeves measurements. We plan to modify the starting position from relative to absolute values or add a calibration period to ensure accurate measurements across all poses by aligning sensor values at 0. Furthermore, when measuring errors from various joints, we noticed that the toe angles consistently demonstrate low error rates. This is likely due to minimal movement in the toe angle during these poses. To better evaluate and compare the model's performance on joints, it would be preferable to use percentage error measuring systems or add poses that include obvious toe movement. 

In addition, we discovered that MoCap systems can encounter occlusion problems, which can affect the accuracy of ground truth labels in our dataset. As a result, we plan to thoroughly distill the ground truth data and ensure that as supervised information, MoCap feedback is accurately interpreted to achieve improved accuracy in the future. What's more, we have mentioned using Fourier resampling to make the recordings from wearable knee sleeves more smooth in \autoref{sec:dataset}, but Fourier transform can't thoroughly solve the uneven time interval problem, which may cause drifted predictions in the test as in \autoref{fig:squat_bluetooth}. Future methods will be recommended to include more specific and complex algorithms focusing on lost time points to address Bluetooth issues. 

\section{Conclusions}
\label{sec:conclusion}
We provide a comprehensive dataset and baseline model for 3D human pose estimation with a pair of durable, stretchable, wearable sensors. We demonstrate our ML model pipeline's effectiveness across various scenarios including generalization to unseen tasks and individuals. We collected a synchronized dataset that comprised time-series data from our smart Knee Sleeves and corresponding ground truth labels from MoCap system. By utilizing these perception outcomes as guidance, our system was able to generate 3D human models solely based on the wearable sensor-integrative apparel readings of individuals performing diverse activities. We achieved an average RMSE of 7.21 degrees across eight joints in the lower body compared to commercially available MoCap tools. Our work offers a novel sensing modality that complements traditional vision systems and enables human pose estimation without being impacted by visual obstructions in a seamless and confidential manner. This innovation has potential applications from home fitness to sports analysis, personalized healthcare, and physical rehabilitation focusing on pose and movement estimation.

\begin{ack}
The smart Knee Sleeves and related app and software for data readout are provided by Texavie Technologies Inc. Texavie collects all wearable sensor data we analyzed in this paper. We express our gratitude to the volunteers who participated in the data collection experiment, as well as to the anonymous reviewers for their valuable comments and discussions. This work received partial support from the University of British Columbia. The opinions, findings, conclusions, and recommendations presented in this paper belong to the authors and do not necessarily represent the views of the funding agencies or the government. The authors would like to thank the useful discussion with Hao Zhang.
\end{ack}

\bibliographystyle{unsrt}
\bibliography{Ref}

\begin{thebibliography}{1}

\bibitem{huynh2009metrics}
Du~Q Huynh.
\newblock Metrics for 3d rotations: Comparison and analysis.
\newblock {\em Journal of Mathematical Imaging and Vision}, 35:155--164, 2009.

\bibitem{luo2021learning}
Yiyue Luo, Yunzhu Li, Pratyusha Sharma, Wan Shou, Kui Wu, Michael Foshey, Beichen Li, Tom{\'a}s Palacios, Antonio Torralba, and Wojciech Matusik.
\newblock Learning human--environment interactions using conformal tactile textiles.
\newblock {\em Nature Electronics}, 4(3):193--201, 2021.

\bibitem{delpreto2022actionsense}
Joseph DelPreto, Chao Liu, Yiyue Luo, Michael Foshey, Yunzhu Li, Antonio Torralba, Wojciech Matusik, and Daniela Rus.
\newblock Actionsense: A multimodal dataset and recording framework for human activities using wearable sensors in a kitchen environment.
\newblock {\em Advances in Neural Information Processing Systems}, 35:13800--13813, 2022.

\bibitem{luo2021intelligent}
Yiyue Luo, Yunzhu Li, Michael Foshey, Wan Shou, Pratyusha Sharma, Tom{\'a}s Palacios, Antonio Torralba, and Wojciech Matusik.
\newblock Intelligent carpet: Inferring 3d human pose from tactile signals.
\newblock In {\em Proceedings of the IEEE/CVF Conference on Computer Vision and Pattern Recognition}, pages 11255--11265, 2021.

\bibitem{tan2022predicting}
Jay-Shian Tan, Sawitchaya Tippaya, Tara Binnie, Paul Davey, Kathryn Napier, JP~Caneiro, Peter Kent, Anne Smith, Peter O’Sullivan, and Amity Campbell.
\newblock Predicting knee joint kinematics from wearable sensor data in people with knee osteoarthritis and clinical considerations for future machine learning models.
\newblock {\em Sensors}, 22(2):446, 2022.

\bibitem{huang2018deep}
Yinghao Huang, Manuel Kaufmann, Emre Aksan, Michael~J Black, Otmar Hilliges, and Gerard Pons-Moll.
\newblock Deep inertial poser: Learning to reconstruct human pose from sparse inertial measurements in real time.
\newblock {\em ACM Transactions on Graphics (TOG)}, 37(6):1--15, 2018.

\end{thebibliography}


\begin{thebibliography}{10}

\bibitem{zheng2020deep}
Ce~Zheng, Wenhan Wu, Chen Chen, Taojiannan Yang, Sijie Zhu, Ju~Shen, Nasser Kehtarnavaz, and Mubarak Shah.
\newblock Deep learning-based human pose estimation: A survey.
\newblock {\em arXiv preprint arXiv:2012.13392}, 2020.

\bibitem{li2018pose}
Dangwei Li, Xiaotang Chen, Zhang Zhang, and Kaiqi Huang.
\newblock Pose guided deep model for pedestrian attribute recognition in surveillance scenarios.
\newblock In {\em 2018 IEEE International Conference on Multimedia and Expo (ICME)}, pages 1--6. IEEE, 2018.

\bibitem{singh2018eye}
Amarjot Singh, Devendra Patil, and SN~Omkar.
\newblock Eye in the sky: Real-time drone surveillance system (dss) for violent individuals identification using scatternet hybrid deep learning network.
\newblock In {\em Proceedings of the IEEE Conference on Computer Vision and Pattern Recognition Workshops}, pages 1629--1637, 2018.

\bibitem{zhou2020single}
Zhihao Zhou, Sean Padgett, Zhixiang Cai, Giorgio Conta, Yufen Wu, Qiang He, Songlin Zhang, Chenchen Sun, Jun Liu, Endong Fan, et~al.
\newblock Single-layered ultra-soft washable smart textiles for all-around ballistocardiograph, respiration, and posture monitoring during sleep.
\newblock {\em Biosensors and Bioelectronics}, 155:112064, 2020.

\bibitem{meng2020wireless}
Keyu Meng, Shenlong Zhao, Yihao Zhou, Yufen Wu, Songlin Zhang, Qiang He, Xue Wang, Zhihao Zhou, Wenjing Fan, Xulong Tan, et~al.
\newblock A wireless textile-based sensor system for self-powered personalized health care.
\newblock {\em Matter}, 2(4):896--907, 2020.

\bibitem{ahad2019vision}
Md~Atiqur~Rahman Ahad, Anindya~Das Antar, and Omar Shahid.
\newblock Vision-based action understanding for assistive healthcare: A short review.
\newblock In {\em CVPR Workshops}, pages 1--11, 2019.

\bibitem{lei2019survey}
Qing Lei, Ji-Xiang Du, Hong-Bo Zhang, Shuang Ye, and Duan-Sheng Chen.
\newblock A survey of vision-based human action evaluation methods.
\newblock {\em Sensors}, 19(19):4129, 2019.

\bibitem{maurice2019human}
Pauline Maurice, Adrien Malais{\'e}, Cl{\'e}lie Amiot, Nicolas Paris, Guy-Junior Richard, Olivier Rochel, and Serena Ivaldi.
\newblock Human movement and ergonomics: An industry-oriented dataset for collaborative robotics.
\newblock {\em The International Journal of Robotics Research}, 38(14):1529--1537, 2019.

\bibitem{Chen_2018_CVPR}
Yilun Chen, Zhicheng Wang, Yuxiang Peng, Zhiqiang Zhang, Gang Yu, and Jian Sun.
\newblock Cascaded pyramid network for multi-person pose estimation.
\newblock In {\em Proceedings of the IEEE Conference on Computer Vision and Pattern Recognition (CVPR)}, June 2018.

\bibitem{zhao2018rf}
Mingmin Zhao, Yonglong Tian, Hang Zhao, Mohammad~Abu Alsheikh, Tianhong Li, Rumen Hristov, Zachary Kabelac, Dina Katabi, and Antonio Torralba.
\newblock Rf-based 3d skeletons.
\newblock In {\em Proceedings of the 2018 Conference of the ACM Special Interest Group on Data Communication}, pages 267--281, 2018.

\bibitem{zhao2018through}
Mingmin Zhao, Tianhong Li, Mohammad Abu~Alsheikh, Yonglong Tian, Hang Zhao, Antonio Torralba, and Dina Katabi.
\newblock Through-wall human pose estimation using radio signals.
\newblock In {\em Proceedings of the IEEE Conference on Computer Vision and Pattern Recognition}, pages 7356--7365, 2018.

\bibitem{luo2021intelligent}
Yiyue Luo, Yunzhu Li, Michael Foshey, Wan Shou, Pratyusha Sharma, Tom{\'a}s Palacios, Antonio Torralba, and Wojciech Matusik.
\newblock Intelligent carpet: Inferring 3d human pose from tactile signals.
\newblock In {\em Proceedings of the IEEE/CVF Conference on Computer Vision and Pattern Recognition}, pages 11255--11265, 2021.

\bibitem{luo2021learning}
Yiyue Luo, Yunzhu Li, Pratyusha Sharma, Wan Shou, Kui Wu, Michael Foshey, Beichen Li, Tom{\'a}s Palacios, Antonio Torralba, and Wojciech Matusik.
\newblock Learning human--environment interactions using conformal tactile textiles.
\newblock {\em Nature Electronics}, 4(3):193--201, 2021.

\bibitem{delpreto2022actionsense}
Joseph DelPreto, Chao Liu, Yiyue Luo, Michael Foshey, Yunzhu Li, Antonio Torralba, Wojciech Matusik, and Daniela Rus.
\newblock Actionsense: A multimodal dataset and recording framework for human activities using wearable sensors in a kitchen environment.
\newblock {\em Advances in Neural Information Processing Systems}, 35:13800--13813, 2022.

\bibitem{shi2020deep}
Qiongfeng Shi, Zixuan Zhang, Tianyiyi He, Zhongda Sun, Bingjie Wang, Yuqin Feng, Xuechuan Shan, Budiman Salam, and Chengkuo Lee.
\newblock Deep learning enabled smart mats as a scalable floor monitoring system.
\newblock {\em Nature Communications}, 11(1):4609, 2020.

\bibitem{ghosh2023sports}
Indrajeet Ghosh, Sreenivasan Ramasamy~Ramamurthy, Avijoy Chakma, and Nirmalya Roy.
\newblock Sports analytics review: Artificial intelligence applications, emerging technologies, and algorithmic perspective.
\newblock {\em Wiley Interdisciplinary Reviews: Data Mining and Knowledge Discovery}, page e1496, 2023.

\bibitem{seshadri2019wearable}
Dhruv~R Seshadri, Ryan~T Li, James~E Voos, James~R Rowbottom, Celeste~M Alfes, Christian~A Zorman, and Colin~K Drummond.
\newblock Wearable sensors for monitoring the physiological and biochemical profile of the athlete.
\newblock {\em NPJ Digital Medicine}, 2(1):72, 2019.

\bibitem{wang2021robust}
Dongkai Wang, Shiliang Zhang, and Gang Hua.
\newblock Robust pose estimation in crowded scenes with direct pose-level inference.
\newblock {\em Advances in Neural Information Processing Systems}, 34:6278--6289, 2021.

\bibitem{jin2020whole}
Sheng Jin, Lumin Xu, Jin Xu, Can Wang, Wentao Liu, Chen Qian, Wanli Ouyang, and Ping Luo.
\newblock Whole-body human pose estimation in the wild.
\newblock In {\em Computer Vision--ECCV 2020: 16th European Conference, Glasgow, UK, August 23--28, 2020, Proceedings, Part IX 16}, pages 196--214. Springer, 2020.

\bibitem{martinez2017simple}
Julieta Martinez, Rayat Hossain, Javier Romero, and James~J Little.
\newblock A simple yet effective baseline for 3d human pose estimation.
\newblock In {\em Proceedings of the IEEE International Conference on Computer Vision}, pages 2640--2649, 2017.

\bibitem{li2022mhformer}
Wenhao Li, Hong Liu, Hao Tang, Pichao Wang, and Luc Van~Gool.
\newblock Mhformer: Multi-hypothesis transformer for 3d human pose estimation.
\newblock In {\em Proceedings of the IEEE/CVF Conference on Computer Vision and Pattern Recognition}, pages 13147--13156, 2022.

\bibitem{papandreou2017towards}
George Papandreou, Tyler Zhu, Nori Kanazawa, Alexander Toshev, Jonathan Tompson, Chris Bregler, and Kevin Murphy.
\newblock Towards accurate multi-person pose estimation in the wild.
\newblock In {\em Proceedings of the IEEE Conference on Computer Vision and Pattern Recognition}, pages 4903--4911, 2017.

\bibitem{zhang2021direct}
Jianfeng Zhang, Yujun Cai, Shuicheng Yan, Jiashi Feng, et~al.
\newblock Direct multi-view multi-person 3d pose estimation.
\newblock {\em Advances in Neural Information Processing Systems}, 34:13153--13164, 2021.

\bibitem{gong2021poseaug}
Kehong Gong, Jianfeng Zhang, and Jiashi Feng.
\newblock Poseaug: A differentiable pose augmentation framework for 3d human pose estimation.
\newblock In {\em Proceedings of the IEEE/CVF Conference on Computer Vision and Pattern Recognition}, pages 8575--8584, 2021.

\bibitem{an2022mri}
Sizhe An, Yin Li, and Umit Ogras.
\newblock mri: Multi-modal 3d human pose estimation dataset using mmwave, rgb-d, and inertial sensors.
\newblock {\em arXiv preprint arXiv:2210.08394}, 2022.

\bibitem{chen2015utd}
Chen Chen, Roozbeh Jafari, and Nasser Kehtarnavaz.
\newblock Utd-mhad: A multimodal dataset for human action recognition utilizing a depth camera and a wearable inertial sensor.
\newblock In {\em 2015 IEEE International Conference on Image Processing (ICIP)}, pages 168--172. IEEE, 2015.

\bibitem{kanazawa2018end}
Angjoo Kanazawa, Michael~J Black, David~W Jacobs, and Jitendra Malik.
\newblock End-to-end recovery of human shape and pose.
\newblock In {\em Proceedings of the IEEE Conference on Computer Vision and Pattern Recognition}, pages 7122--7131, 2018.

\bibitem{Osman2020STAR}
Ahmed A.~A. Osman, Timo Bolkart, and Michael~J. Black.
\newblock Star: Sparse trained articulated human body regressor.
\newblock In Andrea Vedaldi, Horst Bischof, Thomas Brox, and Jan-Michael Frahm, editors, {\em Computer Vision -- ECCV 2020}, pages 598--613, Cham, 2020. Springer International Publishing.

\bibitem{loper2015smpl}
Matthew Loper, Naureen Mahmood, Javier Romero, Gerard Pons-Moll, and Michael~J Black.
\newblock Smpl: A skinned multi-person linear model.
\newblock {\em ACM Transactions on Graphics (TOG)}, 34(6):1--16, 2015.

\bibitem{cao2017realtime}
Zhe Cao, Tomas Simon, Shih-En Wei, and Yaser Sheikh.
\newblock Realtime multi-person 2d pose estimation using part affinity fields.
\newblock In {\em Proceedings of the IEEE Conference on Computer Vision and Pattern Recognition}, pages 7291--7299, 2017.

\bibitem{simo2012single}
Edgar Simo-Serra, Arnau Ramisa, Guillem Alenya, Carme Torras, and Francesc Moreno-Noguer.
\newblock Single image 3d human pose estimation from noisy observations.
\newblock In {\em 2012 IEEE Conference on Computer Vision and Pattern Recognition}, pages 2673--2680. IEEE, 2012.

\bibitem{zhang2022voxeltrack}
Yifu Zhang, Chunyu Wang, Xinggang Wang, Wenyu Liu, and Wenjun Zeng.
\newblock Voxeltrack: Multi-person 3d human pose estimation and tracking in the wild.
\newblock {\em IEEE Transactions on Pattern Analysis and Machine Intelligence}, 45(2):2613--2626, 2022.

\bibitem{fang2017rmpe}
Hao-Shu Fang, Shuqin Xie, Yu-Wing Tai, and Cewu Lu.
\newblock Rmpe: Regional multi-person pose estimation.
\newblock In {\em Proceedings of the IEEE International Conference on Computer Vision}, pages 2334--2343, 2017.

\bibitem{wang2022distribution}
Zitian Wang, Xuecheng Nie, Xiaochao Qu, Yunpeng Chen, and Si~Liu.
\newblock Distribution-aware single-stage models for multi-person 3d pose estimation.
\newblock In {\em Proceedings of the IEEE/CVF Conference on Computer Vision and Pattern Recognition}, pages 13096--13105, 2022.

\bibitem{von2017sparse}
Timo Von~Marcard, Bodo Rosenhahn, Michael~J Black, and Gerard Pons-Moll.
\newblock Sparse inertial poser: Automatic 3d human pose estimation from sparse imus.
\newblock In {\em Computer graphics forum}, volume~36, pages 349--360. Wiley Online Library, 2017.

\bibitem{huang2018deep}
Yinghao Huang, Manuel Kaufmann, Emre Aksan, Michael~J Black, Otmar Hilliges, and Gerard Pons-Moll.
\newblock Deep inertial poser: Learning to reconstruct human pose from sparse inertial measurements in real time.
\newblock {\em ACM Transactions on Graphics (TOG)}, 37(6):1--15, 2018.

\bibitem{yi2022physical}
Xinyu Yi, Yuxiao Zhou, Marc Habermann, Soshi Shimada, Vladislav Golyanik, Christian Theobalt, and Feng Xu.
\newblock Physical inertial poser (pip): Physics-aware real-time human motion tracking from sparse inertial sensors.
\newblock In {\em Proceedings of the IEEE/CVF Conference on Computer Vision and Pattern Recognition}, pages 13167--13178, 2022.

\bibitem{guzov2021human}
Vladimir Guzov, Aymen Mir, Torsten Sattler, and Gerard Pons-Moll.
\newblock Human poseitioning system (hps): 3d human pose estimation and self-localization in large scenes from body-mounted sensors.
\newblock In {\em Proceedings of the IEEE/CVF Conference on Computer Vision and Pattern Recognition}, pages 4318--4329, 2021.

\bibitem{jiang2022avatarposer}
Jiaxi Jiang, Paul Streli, Huajian Qiu, Andreas Fender, Larissa Laich, Patrick Snape, and Christian Holz.
\newblock Avatarposer: Articulated full-body pose tracking from sparse motion sensing.
\newblock In {\em European Conference on Computer Vision}, pages 443--460. Springer, 2022.

\bibitem{madadi2020smplr}
Meysam Madadi, Hugo Bertiche, and Sergio Escalera.
\newblock Smplr: Deep learning based smpl reverse for 3d human pose and shape recovery.
\newblock {\em Pattern Recognition}, 106:107472, 2020.

\bibitem{liao2020deep}
Yalin Liao, Aleksandar Vakanski, and Min Xian.
\newblock A deep learning framework for assessing physical rehabilitation exercises.
\newblock {\em IEEE Transactions on Neural Systems and Rehabilitation Engineering}, 28(2):468--477, 2020.

\bibitem{pantic2003toward}
Maja Pantic and Leon~JM Rothkrantz.
\newblock Toward an affect-sensitive multimodal human-computer interaction.
\newblock {\em Proceedings of the IEEE}, 91(9):1370--1390, 2003.

\bibitem{jaimes2007multimodal}
Alejandro Jaimes and Nicu Sebe.
\newblock Multimodal human--computer interaction: A survey.
\newblock {\em Computer vision and image understanding}, 108(1-2):116--134, 2007.

\bibitem{wen2020machine}
Feng Wen, Zhongda Sun, Tianyiyi He, Qiongfeng Shi, Minglu Zhu, Zixuan Zhang, Lianhui Li, Ting Zhang, and Chengkuo Lee.
\newblock Machine learning glove using self-powered conductive superhydrophobic triboelectric textile for gesture recognition in vr/ar applications.
\newblock {\em Advanced science}, 7(14):2000261, 2020.

\bibitem{libanori2022smart}
Alberto Libanori, Guorui Chen, Xun Zhao, Yihao Zhou, and Jun Chen.
\newblock Smart textiles for personalized healthcare.
\newblock {\em Nature Electronics}, 5(3):142--156, 2022.

\bibitem{chun2021artificial}
Sungwoo Chun, Jong-Seok Kim, Yongsang Yoo, Youngin Choi, Sung~Jun Jung, Dongpyo Jang, Gwangyeob Lee, Kang-Il Song, Kum~Seok Nam, Inchan Youn, et~al.
\newblock An artificial neural tactile sensing system.
\newblock {\em Nature Electronics}, 4(6):429--438, 2021.

\bibitem{zarate2022computational}
V~Vechev~J Zarate and B~Thomaszewski~O Hilliges.
\newblock Computational design of kinesthetic garments.
\newblock 2022.

\bibitem{luo2021knitui}
Yiyue Luo, Kui Wu, Tom{\'a}s Palacios, and Wojciech Matusik.
\newblock Knitui: Fabricating interactive and sensing textiles with machine knitting.
\newblock In {\em Proceedings of the 2021 CHI Conference on Human Factors in Computing Systems}, pages 1--12, 2021.

\bibitem{luo2022digital}
Yiyue Luo, Kui Wu, Andrew Spielberg, Michael Foshey, Daniela Rus, Tom{\'a}s Palacios, and Wojciech Matusik.
\newblock Digital fabrication of pneumatic actuators with integrated sensing by machine knitting.
\newblock In {\em Proceedings of the 2022 CHI Conference on Human Factors in Computing Systems}, pages 1--13, 2022.

\bibitem{zhang2021dynamic}
Qiang Zhang, Yunzhu Li, Yiyue Luo, Wan Shou, Michael Foshey, Junchi Yan, Joshua~B Tenenbaum, Wojciech Matusik, and Antonio Torralba.
\newblock Dynamic modeling of hand-object interactions via tactile sensing.
\newblock In {\em 2021 IEEE/RSJ International Conference on Intelligent Robots and Systems (IROS)}, pages 2874--2881. IEEE, 2021.

\bibitem{guo2023texonmask}
Zengrong Guo and Rong-Hao Liang.
\newblock Texonmask: Facial expression recognition using textile electrodes on commodity facemasks.
\newblock In {\em Proceedings of the 2023 CHI Conference on Human Factors in Computing Systems}, pages 1--15, 2023.

\bibitem{xu2022smart}
Guanghua Xu, Quan Wan, Wenwu Deng, Tao Guo, and Jingyuan Cheng.
\newblock Smart-sleeve: A wearable textile pressure sensor array for human activity recognition.
\newblock {\em Sensors}, 22(5):1702, 2022.

\bibitem{alam2022smart}
Tanvir Alam, Fadoua Saidane, Abdullah Al~Faisal, Ashaduzzaman Khan, and Gaffar Hossain.
\newblock Smart-textile strain sensor for human joint monitoring.
\newblock {\em Sensors and Actuators A: Physical}, 341:113587, 2022.

\bibitem{sundaram2019learning}
Subramanian Sundaram, Petr Kellnhofer, Yunzhu Li, Jun-Yan Zhu, Antonio Torralba, and Wojciech Matusik.
\newblock Learning the signatures of the human grasp using a scalable tactile glove.
\newblock {\em Nature}, 569(7758):698--702, 2019.

\bibitem{liu2021comprehensive}
Chendong Liu, Yilin Zhang, and Huanyu Zhou.
\newblock A comprehensive study of bluetooth low energy.
\newblock In {\em Journal of Physics: Conference Series}, volume 2093, page 012021. IOP Publishing, 2021.

\bibitem{paszke2019pytorch}
Adam Paszke, Sam Gross, Francisco Massa, Adam Lerer, James Bradbury, Gregory Chanan, Trevor Killeen, Zeming Lin, Natalia Gimelshein, Luca Antiga, et~al.
\newblock Pytorch: An imperative style, high-performance deep learning library.
\newblock {\em Advances in Neural Information Processing Systems}, 32, 2019.

\bibitem{huynh2009metrics}
Du~Q Huynh.
\newblock Metrics for 3d rotations: Comparison and analysis.
\newblock {\em Journal of Mathematical Imaging and Vision}, 35:155--164, 2009.

\end{thebibliography}

\section*{Checklist}

\begin{enumerate}

\item For all authors...
\begin{enumerate}
  \item Do the main claims made in the abstract and introduction accurately reflect the paper's contributions and scope?
   \answerYes{}
  \item Did you describe the limitations of your work?
    \answerYes{See \autoref{sec:limitation}. }
  \item Did you discuss any potential negative societal impacts of your work?
    \answerYes{}
  \item Have you read the ethics review guidelines and ensured that your paper conforms to them?
    \answerYes{}
\end{enumerate}

\item If you are including theoretical results...
\begin{enumerate}
  \item Did you state the full set of assumptions of all theoretical results?
    \answerNA{}
	\item Did you include complete proofs of all theoretical results?
    \answerNA{}
\end{enumerate}

\item If you ran experiments (e.g. for benchmarks)...
\begin{enumerate}
  \item Did you include the code, data, and instructions needed to reproduce the main experimental results (either in the supplemental material or as a URL)?
    \answerYes{git@github.com:Zhang-Wenwen/IntelligentKneeBrace.git}
  \item Did you specify all the training details (e.g., data splits, hyperparameters, how they were chosen)?
    \answerYes{See \autoref{sec:implementation}.}
	\item Did you report error bars (e.g., with respect to the random seed after running experiments multiple times)?
    \answerYes{See \autoref{fig:all_seen}, \autoref{tab:RMSE_degree} and 
\autoref{fig:unseen}.}
	\item Did you include the total amount of compute and the type of resources used (e.g., type of GPUs, internal cluster, or cloud provider)?
    \answerYes{See \autoref{sec:implementation}.}
\end{enumerate}

\item If you are using existing assets (e.g., code, data, models) or curating/releasing new assets...
\begin{enumerate}
  \item If your work uses existing assets, did you cite the creators?
    \answerNA{}
  \item Did you mention the license of the assets?
    \answerYes{}
  \item Did you include any new assets either in the supplemental material or as a URL?
    \answerYes{Please refer to \hyperlink{here}{https://github.com/Zhang-Wenwen/IntelligentKneeBrace}.}
  \item Did you discuss whether and how consent was obtained from people whose data you're using/curating?
    \answerYes{Please see \autoref{sec:ethical}.}
  \item Did you discuss whether the data you are using/curating contains personally identifiable information or offensive content?
    \answerYes{Personal information was confidential, and data was used only for research.}
\end{enumerate}

\item If you used crowdsourcing or conducted research with human subjects...
\begin{enumerate}
  \item Did you include the full text of instructions given to participants and screenshots, if applicable?
    \answerYes{}
  \item Did you describe any potential participant risks, with links to Institutional Review Board (IRB) approvals, if applicable?
    \answerYes{Please see \autoref{sec:ethical}.}
  \item Did you include the estimated hourly wage paid to participants and the total amount spent on participant compensation?
    \answerNA{No wages are delivered specifically for the data collection. All participants are volunteering for free of reward. }
\end{enumerate}

\end{enumerate}

\end{document}


\maketitle

\begin{center}
{\Large \textbf{Supplementary Materials}}
\end{center}

This is the Supplementary Material (SI) for our smart Intelligent Knee Sleeves. The SI is formatted as the following: \autoref{apd:apd} provides more details on the analysis of the wearable knee sleeve data. \autoref{eq:Q_d} is the equation we used to quantify the distance between prediction values and ground truth. \autoref{fig:motion_compare} and \autoref{fig:sensor_SI} is the motion capture outputs and corresponding sensor reaction from our stretchable Knee Sleeves under the scenario that some issues happened to the optical system. \autoref{tab:compare} highlights our contribution to related applications. \autoref{tab:multimodal} offers a detailed breakdown of how each modality contributes to the overall performance. \autoref{fig:Label_gen} illustrates the labeling process to obtain the ground truth from a camera-based commercial system. \autoref{fig:squat_bluetooth} shows the time shift in prediction due to the Bluetooth latency issues. \autoref{fig:legRaise_compare} shows the results of unseen tasks in the leg raise pose with seeing 10\% of the unseen task's data. \autoref{apd:data_structure} is the detailed information in the data structure, dimension, and accessibility. \autoref{tab:data_structure} shows the specific dimension and data we input to the model for training. \autoref{tab:dataset_details} is the number of sessions for each pose and each subject on each day. \autoref{tab:exercise_list} summarizes the list of different exercises we conducted during the data collection process. 

Please access the data at \href{https://feel.ece.ubc.ca/smartkneesleeve/}{https://feel.ece.ubc.ca/smartkneesleeve/}. All the data, code, and instructions are stored and can be accessed online in long-term storage repositories. Our work is published under \href{https://www.gnu.org/licenses/gpl-3.0.en.html}{GNU General Public License v3.0}.

\appendix
\renewcommand\thefigure{\Alph{section}\arabic{figure}} 
\setcounter{figure}{0}  
\renewcommand\thetable{\Alph{section}\arabic{table}}
\setcounter{table}{0}
\section{Dataset Analysis}\label{apd:apd}

\autoref{eq:Q_d} is the quaternion distance  \cite{huynh2009metrics} that we used to quantify the model performance on joints' angel predictions. The $D(q_{pred},q_{grd})$ is the normalized quaternions presenting the motion of joints of prediction results and ground truth values. 

\begin{equation}
\left\{
\begin{aligned}
& D(q_{pred},q_{grd}) = 1 -{<q_{pred},q_{grd}>}^2 \\
&<q_{pred},q_{grd}> = <a_1a_2+b_1b_2+c_1c_2+d_1d_2>  \\
& q_{pred} = a_1x+b_1y+c_1z+d_1w \\
& q_{grd} = a_2x+b_2y+c_2z+d_2w  \\
& a_1^2 + b_1^2 + c_1^2 +d_1^2 = 1 \\
& a_2^2 + b_2^2 + c_2^2 +d_2^2 = 1 
\end{aligned}
\right.
\label{eq:Q_d}
\end{equation}

\begin{figure}[htbp]
  \caption{ \textbf{Quaternion example from MoCap system of squatting exercise.} The environment can significantly impact the performance of the motion capture system, as shown in this illustration. Panel (a) presents ground truth values obtained from a motion capture system that functions smoothly without any hindrances, while panel (b) shows ground truth measurements obtained from the motion capture system that faces various challenges, such as occlusion. The corresponding sensor reaction of  the panel (b) is displayed in the \autoref{fig:sensor_SI}}
  \label{fig:motion_compare}
  {\includegraphics[width=1\linewidth]{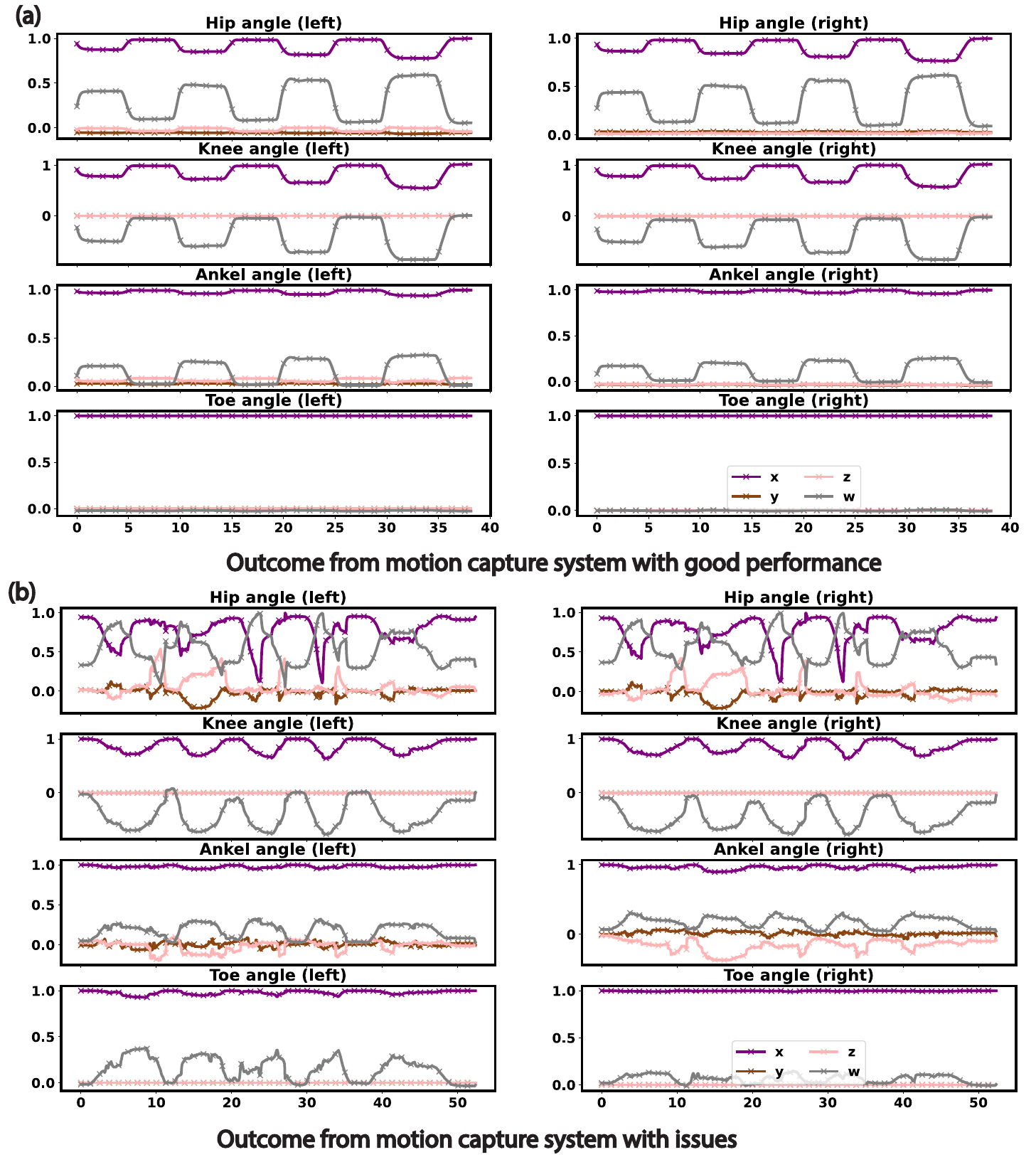}}
\end{figure}

\begin{figure}[t]
  \caption{ \textbf{Normalized sensor reaction to the level of pressure applied}. Pressure sensor response of the exercise displayed in \autoref{fig:motion_compare} (b). The strain reaction from the smart Knee Sleeve matches perfectly with the performed exercise without being affected by any environmental issues. }
  \label{fig:sensor_SI}
  {\includegraphics[width=1\linewidth]{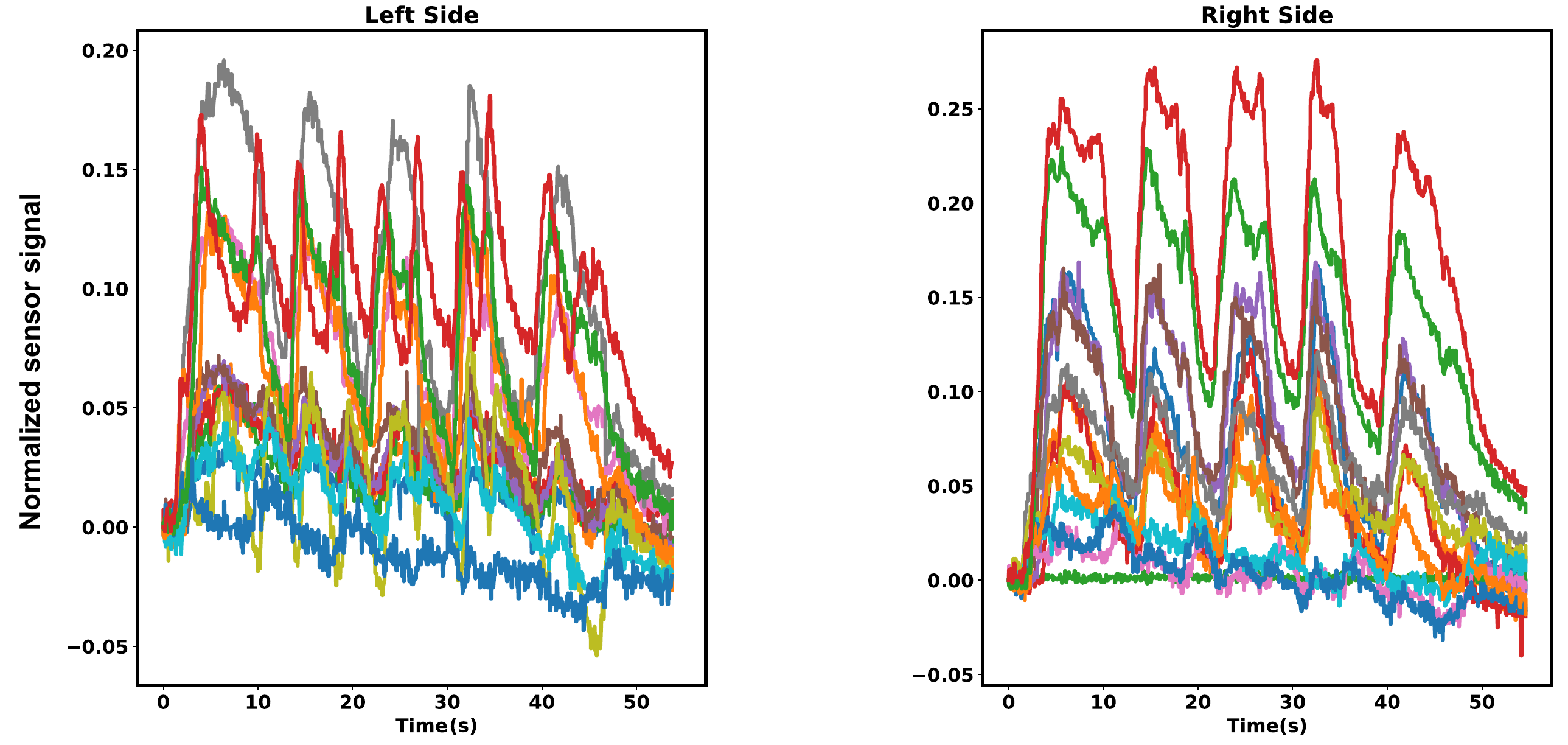}}
\end{figure}

Under adverse environmental conditions, the MoCap system is susceptible to disturbances caused by ambient noise, leading to unstable output. We observed lost and crashed data in the data collection process, as is shown in the \autoref{fig:motion_compare}. Both \autoref{fig:motion_compare} (a) and (b) are the quaternion outcome from the MoCap system during squatting exercise. Panel (a) demonstrates distinct and coordinated joint motion during the exercise, whereas (b) exhibits a higher level of noise and lacks clear depictions of the quaternion alteration throughout the exercise. The corresponding sensor reaction extracted from our wearable Knee Sleeves after data pre-processing is displayed in \autoref{fig:sensor_SI}. The sensor reaction agrees with the squatting exercise, displaying similar patterns as shown in the quaternions illustrated in \autoref{fig:motion_compare} (a). This validates the failure of the MoCap system to record the joint movements in the exercise during certain environmental conditions. When including labels from MoCap systems like \autoref{fig:motion_compare} (b) as supervised information in the training, the model training will be affected by less accurate parameters. Conversely, when employing the data represented in \autoref{fig:motion_compare} (b) as part of the testing set to assess the effectiveness of our baseline model, inaccuracies in the ground truth values will result in higher reported errors compared to the actual values.

\begin{table}[]
\begin{threeparttable}[b]
\caption{\textbf{Comparison table of related work.}}
\label{tab:compare}
\setlength{\tabcolsep}{0.1mm}{
\begin{tabular}{|c|c|c|c|c|c|c|}
\hline
 & \begin{tabular}[c]{@{}c@{}}Sensors\\  Type\end{tabular} & \begin{tabular}[c]{@{}c@{}}Integrated \\ Device\end{tabular} & \begin{tabular}[c]{@{}c@{}}Wireless \\ Steaming\end{tabular} & Task & \begin{tabular}[c]{@{}c@{}}Multi-person \\ scenario issue\end{tabular} & \begin{tabular}[c]{@{}c@{}}Avg \\ RMSE\end{tabular} \\ \hline
\textbf{Our work} & \begin{tabular}[c]{@{}c@{}}IMUs, \\ textile\end{tabular} & y & y & \begin{tabular}[c]{@{}c@{}}Joint orientation \\ inference\end{tabular} & n & \begin{tabular}[c]{@{}c@{}}7.21 deg \\ (Avg angle \\ error)\end{tabular} \\ \hline
\begin{tabular}[c]{@{}c@{}}Luo et al. (2021)\\ \cite{luo2021learning}\end{tabular} & Textile & y & n & \begin{tabular}[c]{@{}c@{}}Pose \\ classification\end{tabular} & n & na \\ \hline
\begin{tabular}[c]{@{}c@{}}DelPreto et al.\\ (2022) \cite{delpreto2022actionsense}\end{tabular} & \begin{tabular}[c]{@{}c@{}}IMU,EMG\\  tactile,\\ camera\end{tabular} & n & n & \begin{tabular}[c]{@{}c@{}}Activity \\ classification \\ in ketch\end{tabular} & n & na \\ \hline
\begin{tabular}[c]{@{}c@{}}Luo et al.\\ (2021) \cite{luo2021intelligent}\end{tabular} & Tactile & y & n & \begin{tabular}[c]{@{}c@{}}21 keypoint \\ estimation, \\ activity\\ classification\end{tabular} & y & \begin{tabular}[c]{@{}c@{}}6.9 cm \\ (Avg location \\ error)\end{tabular} \\ \hline
\begin{tabular}[c]{@{}c@{}}Tan et al.\\ (2022) \cite{tan2022predicting}\end{tabular} & IMUs & n & nm & \begin{tabular}[c]{@{}c@{}}knee flexion/extension \\ prediction\end{tabular} & n & \begin{tabular}[c]{@{}c@{}}9.52 deg \\ (Avg angle\\  error)\end{tabular} \\ \hline
\begin{tabular}[c]{@{}c@{}}Huang et al.\\ (2018) \cite{huang2018deep}\end{tabular} & IMUs & n & nm & Pose estimation & n & \begin{tabular}[c]{@{}c@{}}15.84 deg \\ (Avg angle \\ error)\end{tabular} \\ \hline
\end{tabular}}
\begin{tablenotes}
     \item[*] na: not applicable; nm: not mentioned.
   \end{tablenotes}
   \end{threeparttable}
\end{table}

\begin{table}[]
\caption{\textbf{RMSE in degree unit for smart Knee Sleeve performance.} This supplementary materials provide detailed information on how the multimodal data contributes to the model performance. When training with synthesized data from both IMUs and pressure sensors, we achieved the highest accuracy in the prediction task.}
\label{tab:multimodal}
\begin{tabular}{|c|c|c|c|c|c|c|c|c|}
\hline
\textbf{Training Data} & \textbf{LHip} & \textbf{LKnee} & \textbf{LAnkel} & \textbf{LToe} & \textbf{RHip} & \textbf{RKnee} & \textbf{RAnkel} & \textbf{RToe} \\ \hline
\begin{tabular}[c]{@{}c@{}}IMUs\\ Pressure sensor\end{tabular} & 9.03 & 11.8 & 6.23 & 3.81 & 9.31 & 7.69 & 7.04 & 2.77 \\ \hline
Pressure sensor & 14.06 & 15.76 & 15.60 & 4.80 & 13.32 & 14.54 & 8.01 & 5.49 \\ \hline
IMUs & 11.76 & 11.17 & 14.80 & 4.79 & 10.85 & 11.29 & 6.63 & 5.54 \\ \hline
\end{tabular}
\end{table}

Our Smart Knee Sleeve is a stretchable, user-friendly device suitable for long-term outdoor use. We compare our device with a range of wearable devices, highlighting our innovations in \autoref{tab:compare}, which includes comparisons with works using only IMUs, textile, and sensor systems fused with IMUs and other flexible sensors. We emphasize our device's simplified setup, cost-effectiveness, and robustness, and its potential for more challenging tasks such as dancing and home fitness.

In our study, we utilize both IMUs and pressure sensors to enhance the accuracy of data measurement and reliability of pose estimation tasks. IMUs, while effective, may experience drift caused by minor errors in acceleration data that accumulate over time. Conversely, pressure sensors are highly sensitive to the kinesthetic feedback of joints, even with micro-movements. They can directly detect local deformations, including stretches and pressures of the skin and muscle tissues, caused by joint movements. The fusion of data from IMUs and pressure sensors in our algorithm provides a more comprehensive understanding of the poses. A crucial factor to consider is that IMUs located around the knee joints alone are insufficient to support the predictions of other joints. The pressure sensor we employ can detect muscle activities around the thigh and shank, thereby supporting the prediction of nearby joint activity, such as the hip and ankle. Furthermore, both IMUs and pressure sensors are reliable sources of kinesthetic information under a wide range of conditions, including outdoor, low-light, noisy, and cluttered environments. For a direct comparison demonstrating how the combination of IMU and pressure sensor data results in superior prediction accuracy, We carried out an experiment demonstrating that by integrating multimodal data from both IMUs and pressure sensors, we achieve enhanced accuracy in \autoref{tab:multimodal}.

The latency and shift in predictions caused by Bluetooth and communication systems are illustrated in the \autoref{fig:squat_bluetooth}. The utilization of Bluetooth technology can introduce irregular intervals in wearable recordings, consequently impacting the precision of predictions due to discrepancies between the data extracted from our flexible electronics and the ground truth values from MoCap. The comparison of predictions and ground truth data in \autoref{fig:squat_bluetooth} shows that while the estimated amplitude and frequency of quaternion alteration match well with the ground truth data, there is a noticeable phase shift over time that becomes more severe as time passes by. 

\begin{table}[bth]
\caption{\textbf{RMSE in degree unit for smart Knee Sleeve performance.} This supplementary materials provide detailed information in the  \autoref{tab:RMSE_degree}, including the separate RMSE errors for squat, hamstring, and leg raise poses for all seen scenarios.}
\label{tab:RMSE_degree_si}
\setlength{\tabcolsep}{1.5mm}{
\begin{tabular}{|c|c|c|c|c|c|c|c|c|c|c|}
\hline
\textbf{Scene} & \textbf{Poses} & \textbf{LHip} & \textbf{LKnee} & \textbf{LAnkel} & \textbf{LToe} & \textbf{RHip} & \textbf{RKnee} & \textbf{RAnkel} & \textbf{RToe} & \textbf{Avg} \\ \hline
\multirow{4}{*}{All\_seen} & Squat & 9.90 & 10.20 & 5.18 & 3.96 & 9.59 & 7.91 & 7.39 & 3.14 & 7.16 \\ \cline{2-11} 
 & \begin{tabular}[c]{@{}c@{}}Hamstring \\ Curl\end{tabular} & 5.59 & 18.21 & 7.01 & 3.40 & 9.59 & 4.41 & 4.86 & 1.19 & 6.78 \\ \cline{2-11} 
 & Leg Raise & 8.26 & 8.09 & 8.97 & 3.57 & 7.45 & 9.55 & 7.65 & 2.30 & 6.98 \\ \cline{2-11} 
 & Avg & 9.03 & 11.8 & 6.23 & 3.81 & 9.31 & 7.69 & 7.04 & 2.77 & 7.21 \\ \hline
\multirow{4}{*}{\begin{tabular}[c]{@{}c@{}}Unseen \\ Tasks\end{tabular}} & BendSquat & 17.5 & 14.20 & 12.30 & 4.25 & 17.90 & 15.10 & 12.10 & 5.12 & 12.31 \\ \cline{2-11} 
 & \begin{tabular}[c]{@{}c@{}}Hamstring \\ Curl\end{tabular} & 12.7 & 18.00 & 6.13 & 2.71 & 12.40 & 16.90 & 6.49 & 4.13 & 9.93 \\ \cline{2-11} 
 & Leg Raise & 10.20 & 19.80 & 9.05 & 2.56 & 9.55 & 16.20 & 9.29 & 5.50 & 10.27 \\ \cline{2-11} 
 & Avg & 12.91 & 18.13 & 9.20 & 3.06 & 12.70 & 16.16 & 9.32 & 5.05 & 10.82 \\ \hline
\end{tabular}}
\end{table}

\begin{figure}[t]
  \caption{\textbf{Overview of the label generation process (Squat pose)}. We obtain the ground truth label by utilizing a camera-based MoCap system that incorporates markers attached to the primary joints of the lower body. The outcome of this system is 16-dimensional quaternion data that accurately represents the motion of the major joints. }
  \label{fig:Label_gen}
  {\includegraphics[width=1\linewidth]{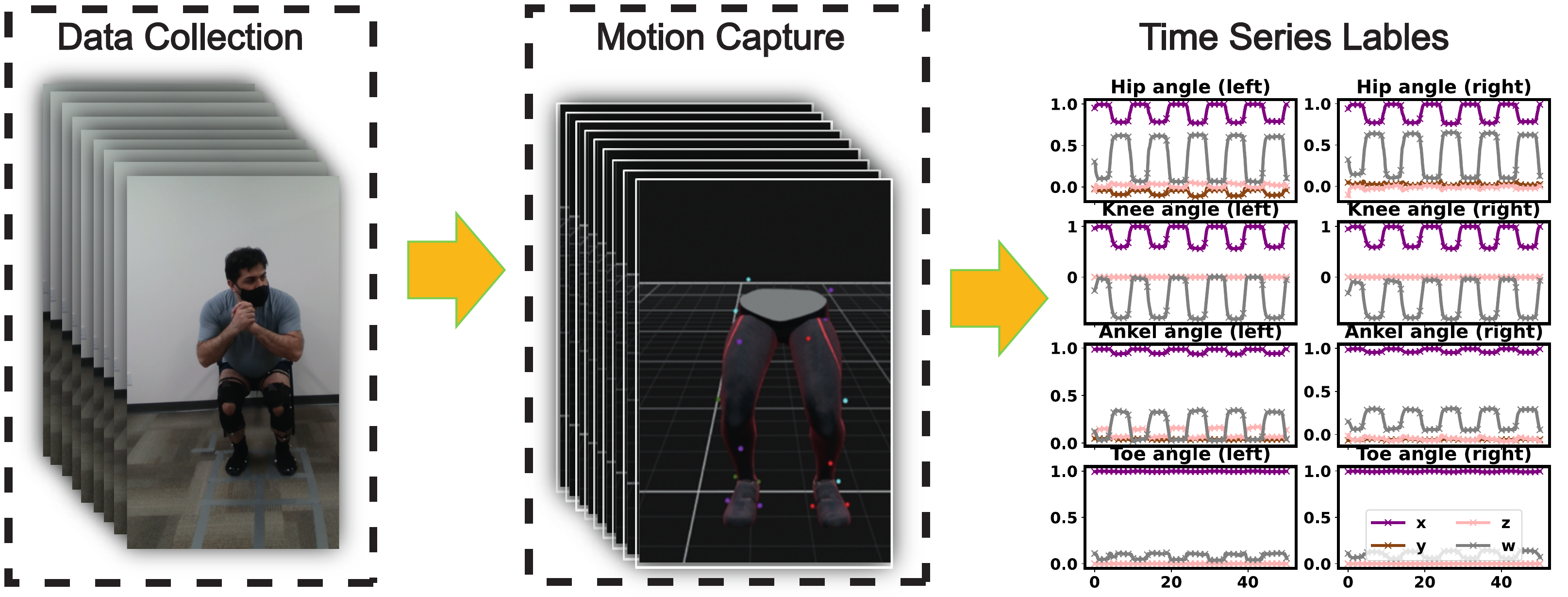}}
\end{figure}

\autoref{fig:Label_gen} illustrates the label generation process as we mentioned in the \autoref{sec:dataset}. We collect the time-series data recording joints movement from the MoCap system as the supervision information for the later training process.

\begin{figure}[t]
  \caption{ \textbf{Quaternion distance comparison from unseen tasks.} When we feed in a limited number of leg raise data (10\%) with setting position as a start point, the outcomes increased compared to the performance that hasn't seen any leg raise data.}
  \label{fig:legRaise_compare}
  {\includegraphics[width=1\linewidth]{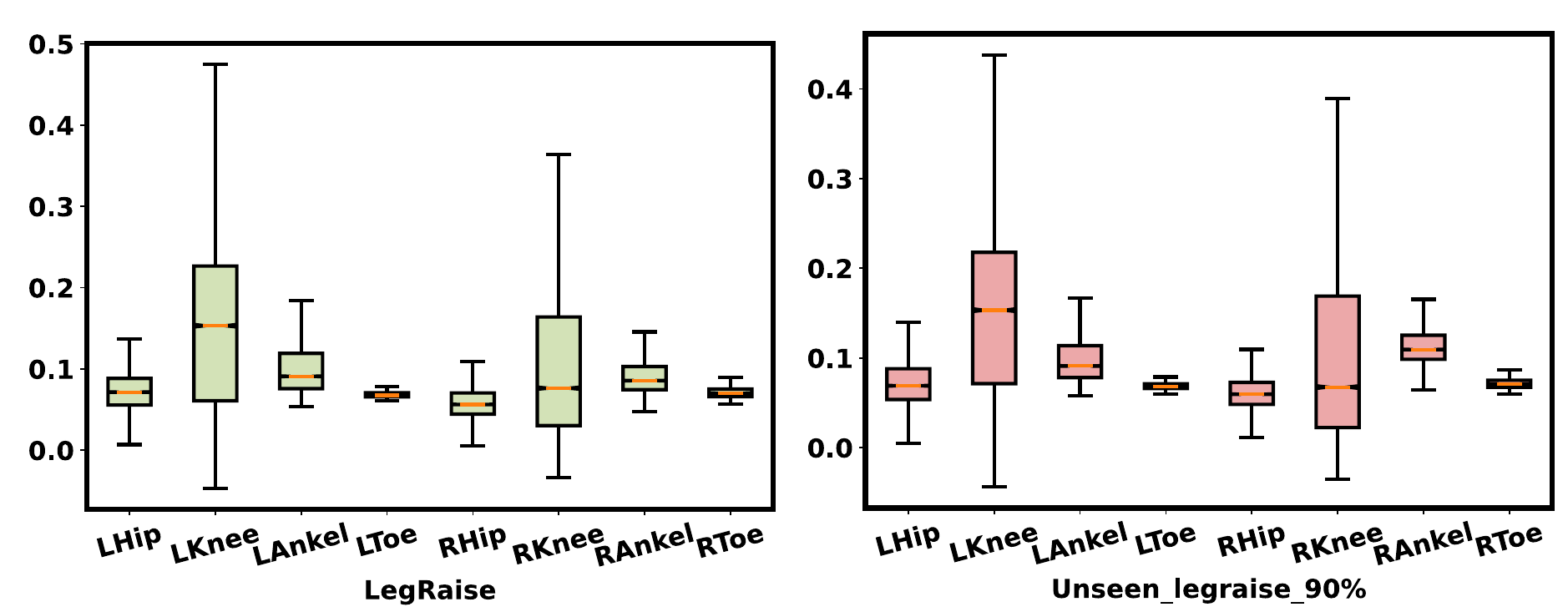}}
\end{figure}

We illustrate a noticeable decline in model performance when dealing with unseen tasks, as shown in in \autoref{fig:unseen}(g-i), and additionally, we perform experiments where the model is exposed to only a restricted portion of labels in \autoref{fig:legRaise_compare}. We split 90\% of the leg raise data as testing and let the model see 10\% of leg raise data during the training process. The \autoref{fig:legRaise_compare} (a) is the result of our baseline model that hasn't seen any leg raise data, while \autoref{fig:legRaise_compare} (b) shows the outcome from the model that is trained with 10\% of leg raise data. We can observe an obvious improvement in the major joints such as the knee and thigh ankle for the left and right sides separately after only seeing a small portion of the data that comes from the leg raise pose. 

We summarized the RMSE in degrees for more details in the seen and unseen task as a supplementary for the \autoref{tab:RMSE_degree}. In all seen scenarios, Bend squat may only take a small portion in the test set due to random sampling, we use overall squatting pose for error calculation here. The joint predictions for different poses display close values in all seen scenarios. 

We provide videos with 3D human models from MoCap ground truth values and our smart Knee Sleeve recordings for comparison in the attachment. The pink dummy in the video is reconstructed from the MoCap values and serves as the ground truth in the video. The blue figure is restored from our smart Knee Sleeve predictions. We provide examples for squatting, hamstring curl, and leg raise poses separately. In the squatting poses, we present RGB camera views, MoCap software recordings, and 3D human model predictions for reference. While the ground truth and our model prediction comparisons are provided for leg raise and hamstring curl poses. The evident correspondence between the ground truth and prediction reconstructions of 3D human model visualizations provides validation for the efficacy of our hardware and model in accurately tracking human activities. 


\begin{figure}[t]
  \caption{ \textbf{Quaternion comparison from ground truth and predictions in time unit (second).} Due to the Bluetooth issue, the predictions are subject to shifting caused by wireless communication latency, and this shift becomes progressively more severe over time.}
  \label{fig:squat_bluetooth}
  {\includegraphics[width=1\linewidth]{Figure/D0_P0_FollowSquat'.pdf}}
\end{figure}

\begin{table}[]
\caption{Mean Squared Derivative results for quaternion predictions on all orientations}
\label{tab:smooth}
\begin{tabular}{|c|c|c|c|c|}
\hline
Orientation & x        & y        & z        & w        \\ \hline
MSD         & 6.92e-05 & 9.82e-09 & 7.08e-09 & 5.46e-04 \\ \hline
\end{tabular}
\end{table}

\begin{figure}[t]
  \caption{Derivative for quaternion predictions on all orientations.}
  \label{fig:deri}
  {\includegraphics[width=1\linewidth]{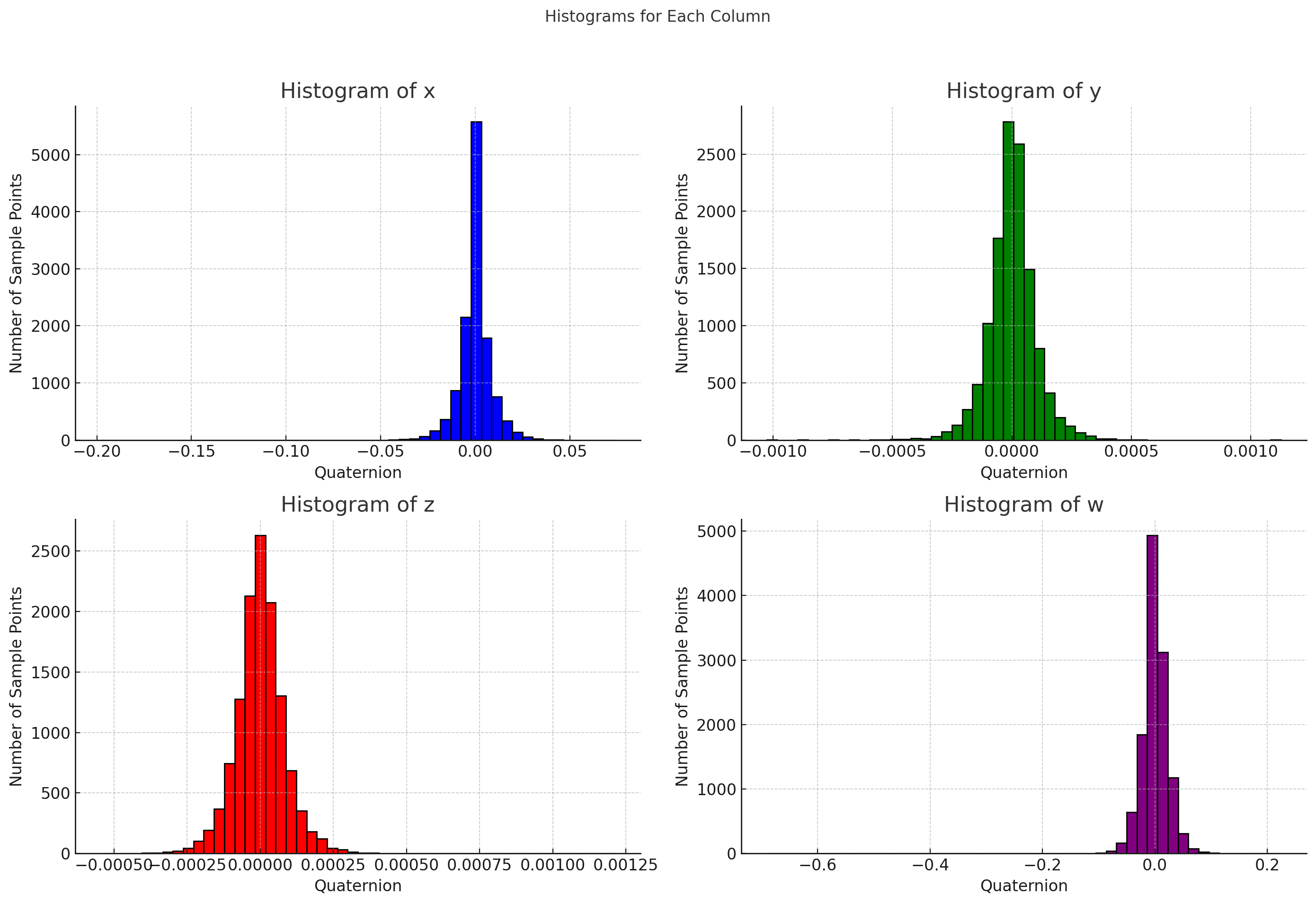}}
\end{figure}

We provide the Mean Squared Derivative (MSD) and derivative of prediction quaternions in \autoref{tab:smooth} and \autoref{fig:deri} to measure the smoothness of our model performance. Our prediction data are quaternions scaled from 0-1, and the MSD for each orientation is at the scale of $10^{-4}$, which proves our prediction results don’t have abrupt changes and are smoothly transitioned.

\section{Dataset Summary and Accessibility}
\label{apd:data_structure}
\textbf{Structure of the dataset}:
Our dataset provides information on synchronized wearable data recordings for smart stretchable Knee Sleeves and camera-based annotations from the MoCap system. The main exercises we focus on are squatting, hamstring curl, and leg raise pose. The structure of the folder is arranged as follows. The data structure is displayed in the \autoref{tab:data_structure}. We have a total of 32-dimensional data collected from the smart Knee Sleeves. The output is 21-dimensional data representing the quaternions of major joints on the lower body.

\begin{itemize}
\item dataset/
\begin{itemize}
\item \texttt{D0/-D11/}
\item \texttt{This folder contains the data files collected from various days. Since each day includes independent calibration of the MoCap system, this may bring in extra noise}.
    \begin{itemize}
      \item \texttt{P0/-P5/}
      \item \texttt{ In this directory, we include files collected on the same day from various subjects. It's worth noting that different participants wore knee sleeves of varying sizes on that particular day.}
       \item \texttt{This folder contains three types of files:}
        \begin{itemize}
              \item \texttt{Exercise.csv. Example: BendSquat.csv}
              \item \texttt{Exercise\_l.csv. Example: BendSquat\_l.csv}
              \item \texttt{Exercise\_r.csv. Example: BendSquat\_r.csv}
              \item \texttt{Exercise.csv files are ground truth annotations collected from the motion capture system, while Exercise\_l.csv and Exercise\_r.csv are readouts from the smart Knee Sleeves of the left and right sides of the body separately. The data structure is displayed in \autoref{tab:data_structure}}
      \end{itemize}
    \end{itemize}
\end{itemize}
\end{itemize}

The summary of \autoref{tab:dataset_details} contains the content of the number of tasks performed by each subject and the number of subjects collected each day. Squatting exercises usually are comprised of several different types of tasks, but we term them as squat together in this table. Please refer to the \autoref{tab:exercise_list} for details of a full list of various exercises we have conducted. In \autoref{tab:exercise_list}, Exercise column is the name of the exercise we guide the participant to do during the data collection process.  \#Total-files column is the total number of files we have with each corresponding exercise, which contains Mocap ground truth data, and the smart Knee Sleeve data for the left and right sides separately. We have one missing side file for the hamstring pose due to battery issues. \#Session column summarizes the sessions we have for each exercise. 

\begin{table}[htbp]
\caption{Data Structure for the wearable benchmarks used in this paper. The last quaternion columns are the relative values calculated from $Quat_0$ and $Quat_1$ attached to the thigh and shank separately.}
\setlength{\tabcolsep}{1.5mm}{
{\begin{tabular}{c|c|c|c|c|c|c|c|c|c}
\toprule
Data      & Time & Pressure Sensor Pins & Acc0 & Quat0 & Gyro0 & Acc1 & Quat1 & Gyro1 & Quat \\    \midrule
Dimension & 1    & 14                 & 3    & 4     & 3     & 3    & 4     & 4     & 4    \\   \bottomrule
\end{tabular}}}
\label{tab:data_structure}
\end{table}

\begin{table}[htbp]
\caption{Details about our smart Knee Sleeve dataset including data collection dates, number of subjects, and tasks. Squatting tasks usually contain multiple categories of tasks, but we refer to them as squatting in general. Please refer to \autoref{tab:exercise_list} for a full list of exercises. }
{\begin{tabular}{|c|c|ccc|c|}
\toprule
\multirow{2}{*}{Date} & \multirow{2}{*}{Subject} & \multicolumn{3}{c|}{Tasks} & \multirow{2}{*}{Summary} \\ \cline{3-5} 
 &  & \multicolumn{1}{c|}{Squat} & \multicolumn{1}{c|}{Hamstring Curl} & Leg Raise &  \\  \midrule
D0 & P0 & \multicolumn{1}{c|}{6} & \multicolumn{1}{c|}{1} & 1 & 8 \\ \hline
D1 & P0 & \multicolumn{1}{c|}{5} & \multicolumn{1}{c|}{1} & 2 & 8 \\ \hline
D2 & P0 & \multicolumn{1}{c|}{7} & \multicolumn{1}{c|}{2} & 2 & 11 \\ \hline
\multirow{2}{*}{D3} & P0 & \multicolumn{1}{c|}{5} & \multicolumn{1}{c|}{1} & 2 & 8 \\ \cline{2-6} 
 & P3 & \multicolumn{1}{c|}{6} & \multicolumn{1}{c|}{0} & 2 & 8 \\ \hline
D4 & P0 & \multicolumn{1}{c|}{7} & \multicolumn{1}{c|}{2} & 2 & 11 \\ \hline
D5 & P0 & \multicolumn{1}{c|}{4} & \multicolumn{1}{c|}{2} & 2 & 8 \\ \hline
\multirow{2}{*}{D6} & P0 & \multicolumn{1}{c|}{1} & \multicolumn{1}{c|}{1} & 0 & 2 \\ \cline{2-6} 
 & P1 & \multicolumn{1}{c|}{2} & \multicolumn{1}{c|}{0} & 2 & 4 \\ \hline
\multirow{2}{*}{D7} & P0 & \multicolumn{1}{c|}{8} & \multicolumn{1}{c|}{2} & 2 & 12 \\ \cline{2-6} 
 & P2 & \multicolumn{1}{c|}{6} & \multicolumn{1}{c|}{1} & 2 & 9 \\ \hline
\multirow{3}{*}{D8} & P0 & \multicolumn{1}{c|}{8} & \multicolumn{1}{c|}{2} & 2 & 12 \\ \cline{2-6} 
 & P1 & \multicolumn{1}{c|}{5} & \multicolumn{1}{c|}{2} & 2 & 9 \\ \cline{2-6} 
 & P4 & \multicolumn{1}{c|}{3} & \multicolumn{1}{c|}{2} & 1 & 6 \\ \hline
\multirow{2}{*}{D9} & P0 & \multicolumn{1}{c|}{5} & \multicolumn{1}{c|}{1} & 2 & 8 \\ \cline{2-6} 
 & P1 & \multicolumn{1}{c|}{8} & \multicolumn{1}{c|}{2} & 2 & 12 \\ \hline
\multirow{2}{*}{D10} & P0 & \multicolumn{1}{c|}{7} & \multicolumn{1}{c|}{2} & 1 & 10 \\ \cline{2-6} 
 & P1 & \multicolumn{1}{c|}{5} & \multicolumn{1}{c|}{2} & 2 & 9 \\ \hline
\multirow{3}{*}{D11} & P0 & \multicolumn{1}{c|}{6} & \multicolumn{1}{c|}{0} & 0 & 6 \\ \cline{2-6} 
 & P1 & \multicolumn{1}{c|}{2} & \multicolumn{1}{c|}{0} & 0 & 2 \\ \cline{2-6} 
 & P5 & \multicolumn{1}{c|}{1} & \multicolumn{1}{c|}{0} & 0 & 1 \\  \hline
\end{tabular}}
\label{tab:dataset_details}
\end{table}

\begin{table}[htbp]
\caption{Detailed list of exercises collected with our smart Knee Sleeve dataset.}
\label{tab:exercise_list}
\begin{tabular}{|c|c|c|}
\hline
Exercise & \# Total files & \# Sessions \\ \hline
AppFastSquat & 3 & 1 \\ \hline
AppSlowSquat & 3 & 1 \\ \hline
FollowSquat & 3 & 1 \\ \hline
HamstringCurlLeft & 44 & 15 \\ \hline
LegRaiseLeft & 48 & 16 \\ \hline
StepwiseSquat & 45 & 15 \\ \hline
TiredSquat & 27 & 9 \\ \hline
ToeOutSquat & 3 & 1 \\ \hline
BendSquat & 48 & 16 \\ \hline
LegRaiseRight & 45 & 15 \\ \hline
SlowSquat & 45 & 15 \\ \hline
Squat & 75 & 25 \\ \hline
WeightSquat & 48 & 16 \\ \hline
HamstringCurlRight & 33 & 11 \\ \hline
StaggeredRightSquat & 9 & 3 \\ \hline
PulseSquat & 6 & 2 \\ \hline
StaggeredLeftSquat & 6 & 2 \\ \hline
Rest-l & 2 & 1 \\ \hline
Rest-r & 2 & 1 \\ \hline
\end{tabular}
\end{table}
\newpage
\section{Ethical Considerations}
\label{sec:ethical}
Participants were fully informed of the study's purpose, potential risks, and benefits. Personal information was kept confidential, and data were used exclusively for research purposes. Our wearable sensors were designed to be safe, minimizing discomfort and skin irritation. We monitored participants to prevent extended use of sensors that could cause skin irritation. We also complied with data protection laws by securely storing all collected data and taking measures to prevent unauthorized access. Our study was conducted ethically and responsibly. The product of our Intelligent Knee sleeves and its related data are the intellectual property of Texavie Technologies Inc. Please refer to Texavie for more details.

\bibliographystyle{unsrt}
\bibliography{Ref}